\pdfoutput=1

\documentclass[11pt]{article}

 \usepackage[final]{acl}

\usepackage{times}
\usepackage{latexsym}
\usepackage{graphicx}
\usepackage{multirow}
\usepackage{arydshln} 
\usepackage{amsmath}
\usepackage{amsfonts}
\usepackage[T1]{fontenc}

\usepackage[utf8]{inputenc}

\usepackage{microtype}
\usepackage{booktabs}
\usepackage{caption}
\usepackage{subfigure}
\newcommand{\m}[1]{\mathbf{#1}}

\newcommand{\tabincell}[2]{\begin{tabular}{@{}#1@{}}#2\end{tabular}}

\def \x{{\bf x}}
\def \y{{\bf y}}
\def \z{{\bf z}}
\def \h{{\bf h}}

\def \q{{\bf q}}
\def \mL{\ell}
\def \mS{{\mathcal S}}
\def \u{{\bf u}}

\title{
Compression  of Generative Pre-trained Language Models via Quantization }

\author{Chaofan Tao$^1$, Lu Hou$^2$, Wei Zhang$^2$, Lifeng Shang$^2$,\\  \textbf{Xin Jiang}$^2$, \textbf{Qun Liu}$^2$, \textbf{Ping Luo}$^1$, \textbf{Ngai Wong}$^1$\\
$^1$The University of Hong Kong \quad
$^2$Huawei Noah’s Ark Lab \\
cftao@connect.hku.hk, pluo@cs.hku.hk, nwong@eee.hku.hk \\
\{houlu3, zhangwei379, shang.lifeng, jiang.xin, qun.liu\}@huawei.com}

\begin{document}
\maketitle
\begin{abstract}
The increasing size of generative Pre-trained Language
Models (PLMs) has greatly increased the demand for model compression.
Despite various methods to compress BERT or its variants, 
there are few attempts to compress generative PLMs, and the underlying difficulty remains unclear.
In this paper, we compress generative PLMs by quantization.
We find that previous quantization methods fail on generative tasks due to the \textit{homogeneous word embeddings} caused by reduced capacity, and  \textit{varied distribution of weights}.
Correspondingly, we propose a token-level contrastive distillation to learn distinguishable word embeddings, 
and a module-wise dynamic scaling to make quantizers adaptive to different modules.
Empirical results on various tasks show that 
our proposed method outperforms the state-of-the-art compression
methods on generative PLMs by a clear margin.  
With comparable performance with the full-precision models, we achieve 14.4$\times$ and 13.4$\times$ compression rates on GPT-2 and BART, respectively.

\end{abstract}

\section{Introduction}
Transformer-based generative pre-trained language models (PLMs)
show strong abilities of multi-task and few-shot learning, 
and achieve remarkable performances on various tasks~\citep{gpt,gpt3,bart,t5,chen2021litegt}. 
However, they are usually expensive in terms of both computation and memory due to a large number of parameters, and the token-by-token generation process. 
Many methods have been proposed to compress PLMs, but mostly focus on understanding tasks like sentence classification with BERT ~\citep{albert,mobilebert,jiao2020tinybert,shen2019q,dynabert}.
Recent works  try to compress GPT-2 using  tensor decomposition 
\citep{edalati2021kronecker}, and knowledge distillation 
\citep{song2020lightpaff}, but the compression ratio achieved
is much smaller than that of BERT.
Yet the underlying 
difficulty remains unclear.

\begin{figure}[t]
	\centering 
	\includegraphics[width=0.45\textwidth]{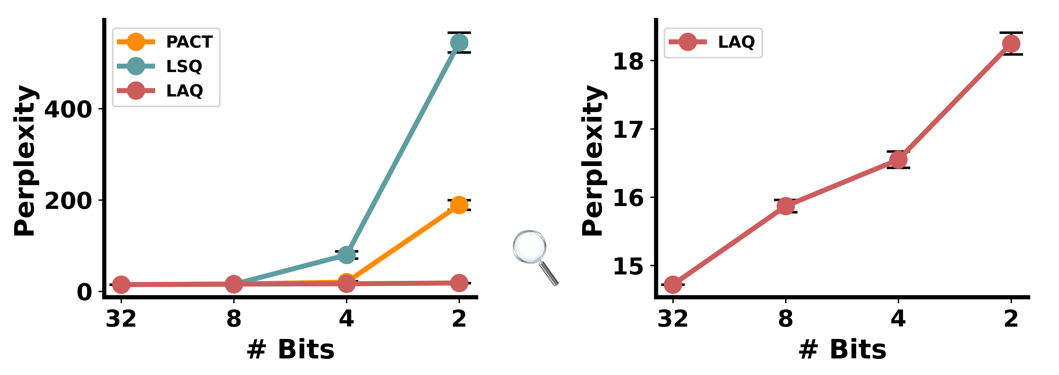}
	\caption{
	Performance of quantized GPT-2 with varying
weight bit-widths and 8-bit activation, using different methods.
	The  right figure takes a closer look at LAQ. }
	
	\label{fig:drop}
			\vspace{-0.1in}
\end{figure}

In this paper, we firstly 
explore compressing generative PLMs by quantizing the parameters from full-precision to lower bits. 
We find that directly applying previous quantization methods designed for BERT or computer vision tasks to generative PLMs lead to poor performance.
Figure~\ref{fig:drop} shows that the performance drops sharply as the 
weight 
bit-width decreases.
To investigate the difficulty of quantizing generative PLMs,
we find that 
the learned embeddings tend to be homogeneous and hard to distinguish due to the reduced capacity caused by quantization, while the weight distributions also vary significantly across different modules and different Transformer layers.
These problems are further magnified due to the nature of sequential left-to-right  prediction of generative PLMs, as the quantization error will accumulate across time. 

To alleviate the above problems, we propose
a token-level contrastive distillation to contrast on tokens and make the word embedding distinguishable.
Besides, we  propose a module-wise dynamic scaling for the quantizer to better adapt to different modules.
Empirical results on language modeling, next utterance prediction and summarization show that compared to the full-precision baseline, 
our quantized GPT and BART (abbreviated as QuantGPT and QuantBART)
achieve comparable performance for 8/4-bit weight, and have only a slight drop for 2-bit weight, while being over 13$\times$ smaller.
QuantGPT also clearly
outperforms previous GPT compression methods on
language modeling. 



To summarize, our main contributions  are: 1) We 
find that generative PLMs are hard to quantize due to 
\textit{homogeneous word embedding} and \textit{varied  weight distribution}. 
2)  We then propose the token-level contrastive distillation and module-wise dynamic scaling, to 
make the word embedding more distinguishable and make quantizers adapt to different modules, respectively. 
3) Empirical results on various tasks show the efficacy of our method.

\section{Difficulty of Qunatizing Generative Pre-trained Language Models}
\label{sec:difficulty}

In this section, we show that it is challenging to train a low-bit generative pre-trained model with conventional quantization approaches directly. 
Before diving into details, we first review the necessary backgrounds of quantization. 

\subsection{Network Quantization}
\label{sec:quantization}
In this paper,
we apply the quantization-aware training~\cite{binaryconnect} 
to generative PLMs.
Specifically,
denote the vectorized full-precision weight as $\m w $,
each forward propagation 
first clips the weight by a positive clipping factor $\alpha$, and then quantizes 
the clipped weight
to $b$-bit as
\begin{equation}
\label{eq:quantize}
\m w_q =   \alpha \cdot Q (\text{clip}(\m w, -\alpha, \alpha)  / \alpha ),
\end{equation}
where $Q$ is the quantization function that maps
each entry in $\text{clip}(\m w, -\alpha, \alpha) / \alpha $ to its closest quantized value in the set of uniform discrete values 
$\{-1,\! -\frac{k-1}{k},\cdots,-\frac{1}{k}, 0,\frac{1}{k}, \cdots, \frac{k-1}{k}, 1\}$ with $k=2^{b-1}-1$.
Then we compute the loss $\ell(\m w_q)$ with $\m w_q$.
During back propagation, we  use the gradient with regard to the quantized weight $\nabla \ell(\m w_q)$ as the Straight-Through-Estimator~\cite{ste} to update full-precision weights $\m w$ 
due to the non-differentiability of $Q(\cdot)$.

A good clipping factor is expected to take 
the majority of full-precision weight into account via clipping, 
\textit{i.e.}, quantizing the range where data are densely distributed to reduce quantization error. 
To solve this problem,  PACT~\citep{choi2018pact} learns a parameterized clipping factor 
and achieves better results than setting a fixed clipping factor.  
Instead of learning the clipping factor, LSQ~\citep{esser2020learned} learns the step size $\alpha/n$, but requires
 a careful initialization and gradient update.

In practice, following previous works on  BERT quantization~\citep{ternarybert,binarybert},
we use layer-wise quantization (\textit{i.e.}, one clipping factor for elements in each weight matrix) for all  weight matrices in the Transformer layers  and row-wise quantization  (\textit{i.e.}, one clipping factor for each word embedding) for the embedding layer.
We use asymmetric uniform quantization for activations after self-attention and GeLU function whose elements are mostly positive, and symmetric uniform quantization for other activations.
We do not quantize layer-normalization layers, skip connections, biases due to small computational overhead.

\subsection{Difficulty Analysis}
\label{expt:difficulty}
We compare the following representative quantization methods  including
 (i) LAQ~\citep{ternarybert} for BERT; 
 (ii) PACT~\citep{choi2018pact}
and  LSQ~\citep{esser2020learned}) 
 for  computer vision tasks,
  to generative pre-trained model,  GPT-2. 
 Figure \ref{fig:drop} shows the performance under
 different weight bit-widths, and
 the performance drops sharply as the bit-width decreases, especially for PACT and LSQ. 
 In the following, we study the potential reasons  behind  the difficulty of  quantizing generative PLMs, by empirically investigating the properties of the word embedding and model parameters.


\paragraph{Homogeneous Word Embedding.}  
\begin{figure*}[htbp]
\centering
	\subfigure[Full-precision.]{
		\includegraphics[width=0.18\textwidth]{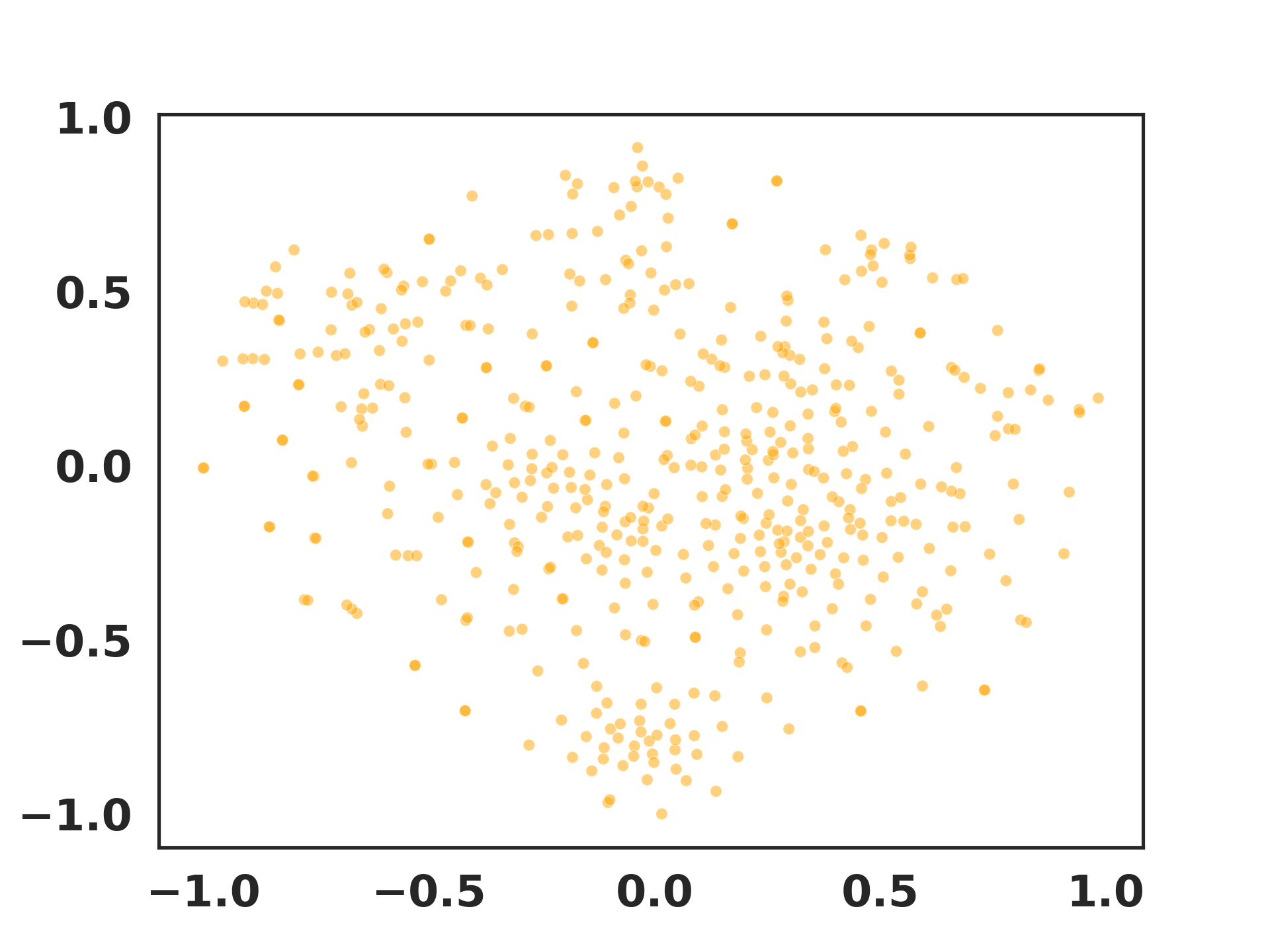}
		\label{fig:train_time}
	}
	\subfigure[PACT.]{
		\includegraphics[width=0.18\textwidth]{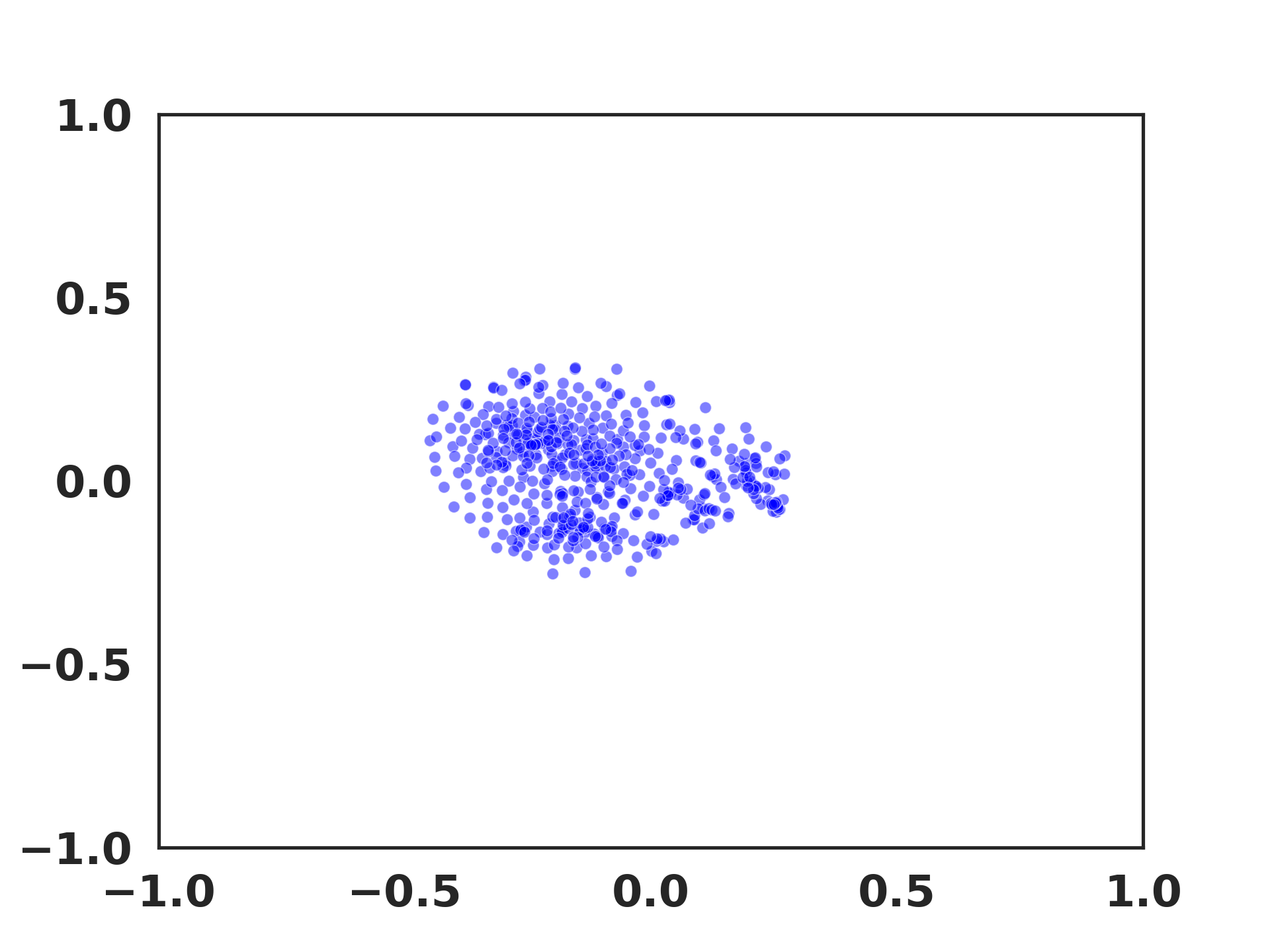}
		\label{fig:PACT_3000}
	}
	\subfigure[LSQ.]{
		\includegraphics[width=0.18\textwidth]{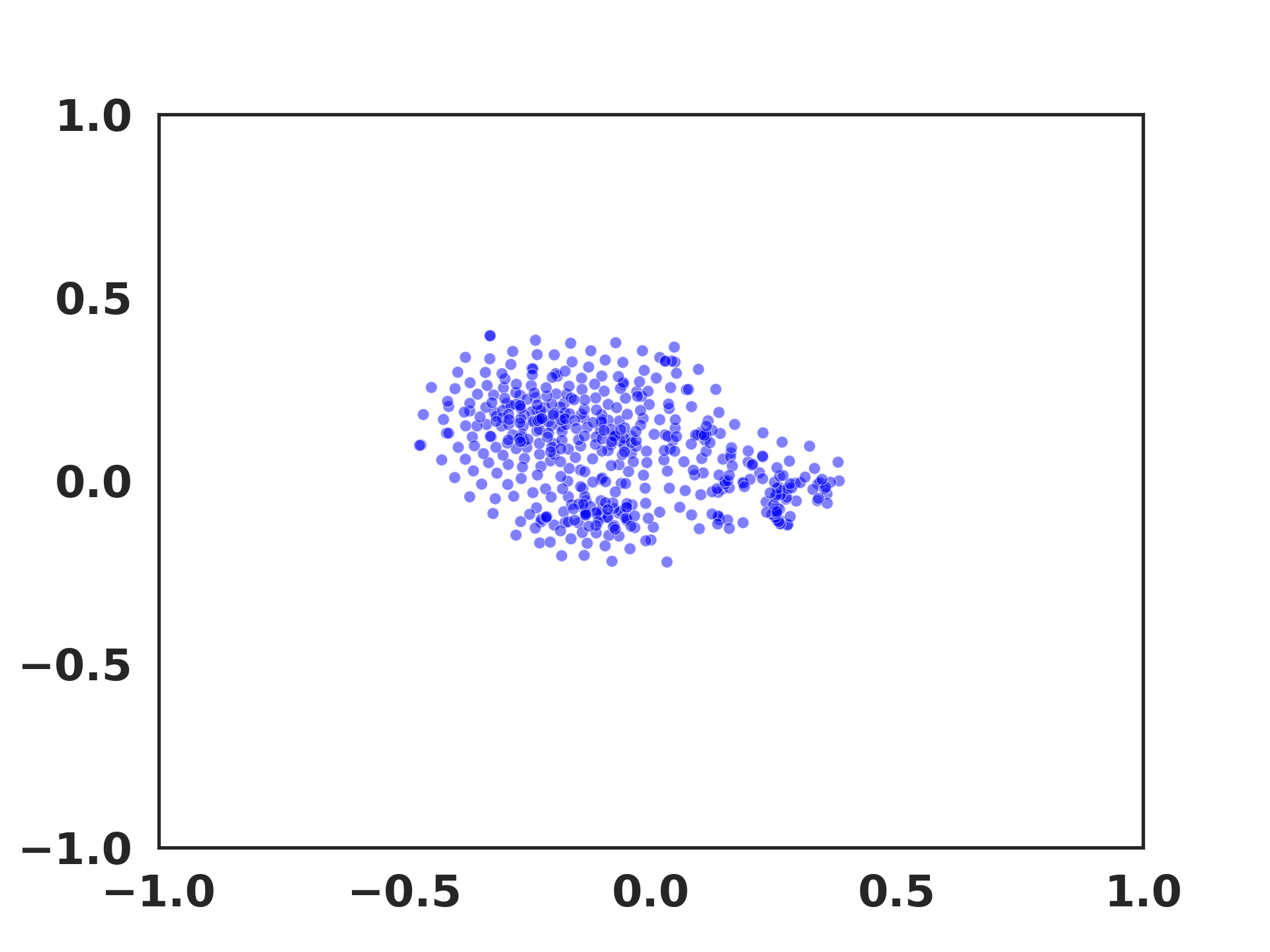}
		\label{fig:lsqq_1000}
	}
	\subfigure[LAQ.]{
		\includegraphics[width=0.18\textwidth]{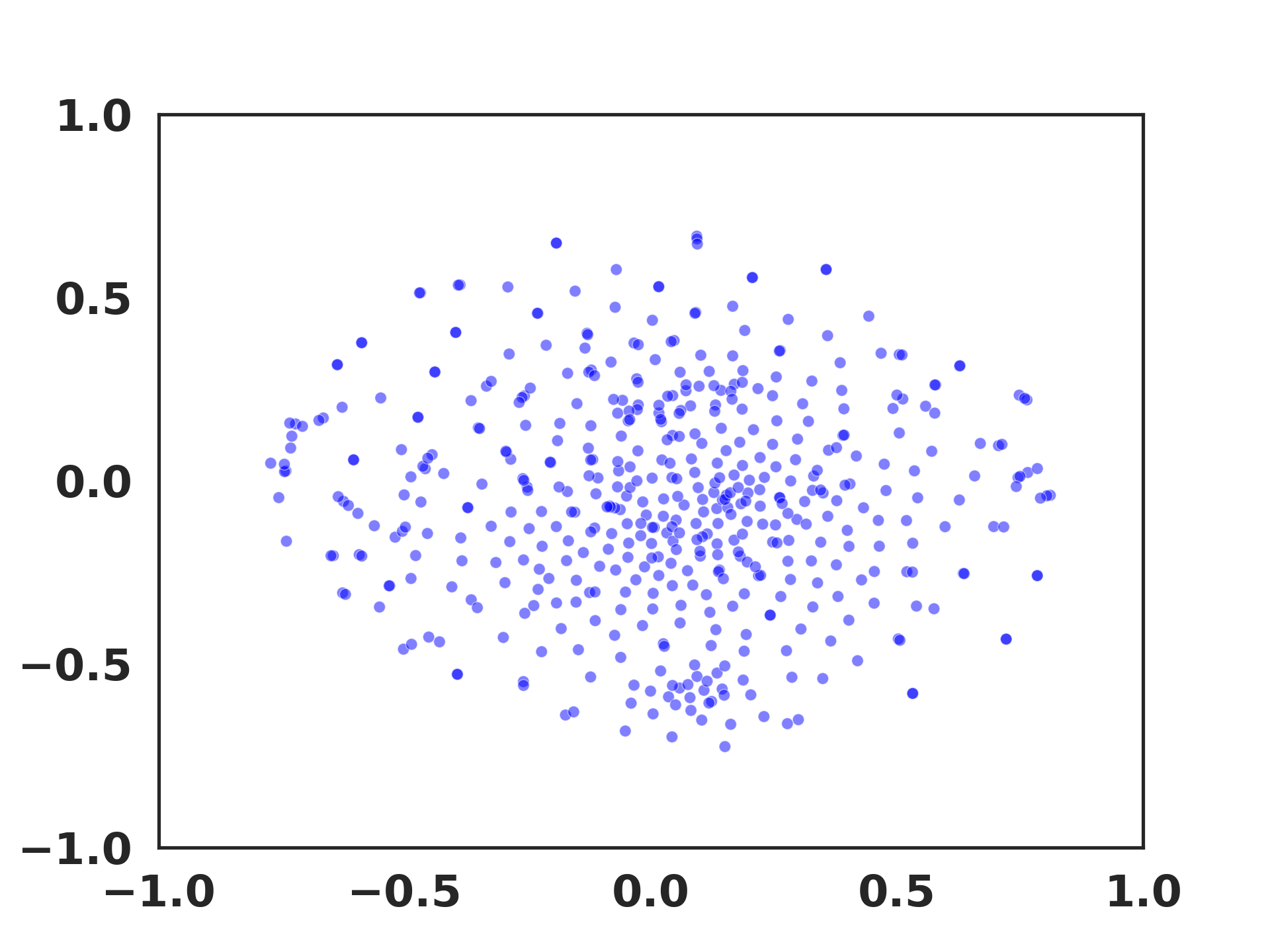}
		\label{fig:laq_2000}
	}
\subfigure[Ours.]{
	\includegraphics[width=0.18\textwidth]{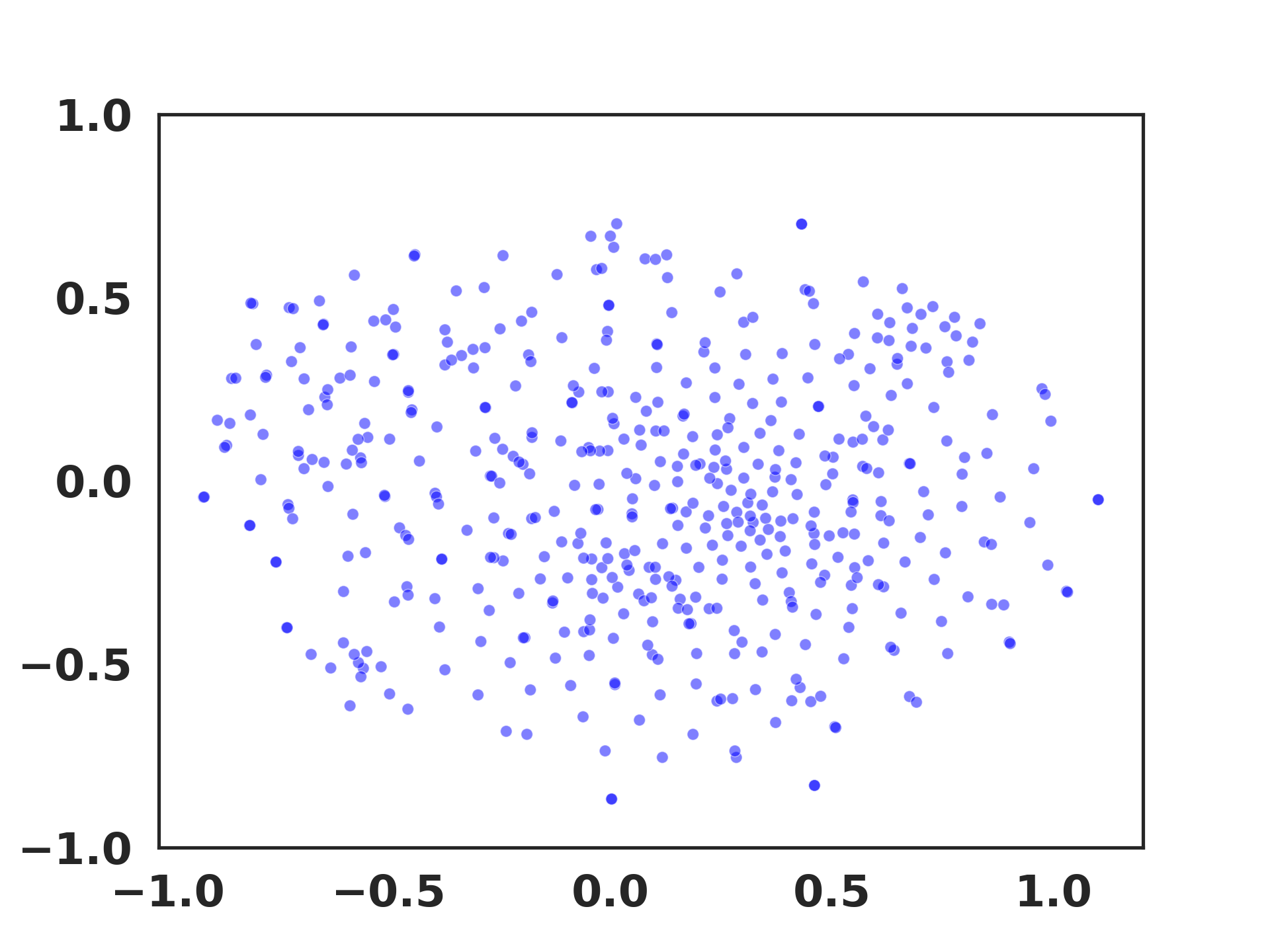}
	\label{fig:ours_ptq}
}
	\vspace{-0.1in}
	\caption{T-SNE visualization  of the most frequent 500 word embeddings, 
	of the full-precision and different 2-bit quantized  models trained on PTB dataset. 
	Embeddings of different methods show different degrees of homogeneity.
	}
	\label{fig:embedding_collapse}
\end{figure*}

\begin{figure*}[htbp]
\centering
 \includegraphics[width=0.95\textwidth]{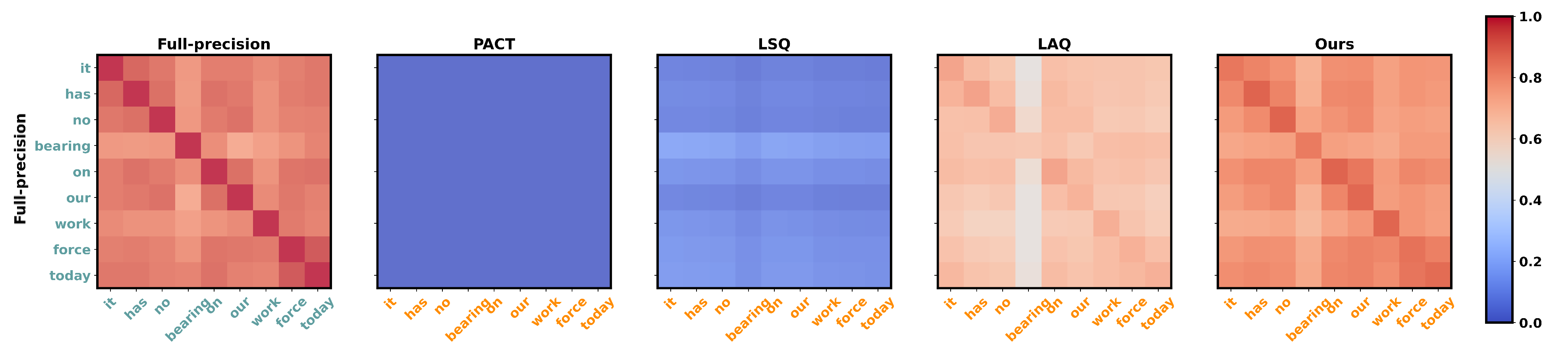}
 \vspace{-0.1in}
\caption{
Matrices representing the cosine similarities between representations of all pairs of tokens in a sentence,
between the full-precision model and 2-bit quantized models trained on PTB dataset.  
Token representations at the last decoder layer of GPT-2 are used.
More visualizations are available in Appendix \ref{apdx:vis_token}.
}
\label{fig_cossim}
\end{figure*}

We first study the difficulty from the learned word embeddings of different models.
In Figure \ref{fig:embedding_collapse}, we  visually compare the distributions of the word embeddings  of 
the full-precision and quantized models under the same scale. 
As can be seen, the word embeddings of the full-precision model are scattered distinguishable,  
while those in previous quantization methods PACT, LSQ and LAQ learn homogeneous word embeddings which
are clustered and less distinguishable, especially for PACT and LSQ. 
We speculate this is caused by the sequential computation nature of GPT.
Specifically, unlike BERT which computes the representation of all tokens in parallel, GPT computes each token in left-to-right order, and the quantization error incurred in the previous tokens will pass on to future tokens, making the learning signal noisier over time, and finally less informative word embeddings. 
 
 A direct consequence of the homogeneous word embedding can be reflected in Figure~\ref{fig_cossim}.
 By comparing Figure~\ref{fig:embedding_collapse} and Figure~\ref{fig_cossim}, 
 we can find that the higher degree of homogeneity in the word embedding of a quantized model, the fewer dependencies among different tokens are kept.
 
As will be discussed in Section~\ref{sec:seq_contrastive_learning}, we propose a token-level contrastive learning to alleviate this problem.
Compared with PACT, LSQ and LAQ, our method 
not only aligns the token representations between the quantized and full-precision networks
(\textit{i.e.}, diagonal boxes), but also captures the dependencies among different tokens (non-diagonal boxes).  
More visualizations are available in Appendix \ref{apdx:vis_token}.
The non-distinguishable word embeddings and poor ability to capture contextualized dependencies
also make methods like PACT and LSQ more likely to 
generate incorrect tokens, \textit{e.g.} 
illogical and repeated text (
Section~\ref{sec:abs_summarization}).


\paragraph{Varied Distribution of Weights.}
\label{section-diverse-weight}
\begin{figure}[htbp]
\centering
	\subfigure[$\m w_o$  at Layer 4.]{
		\includegraphics[width=0.22\textwidth]{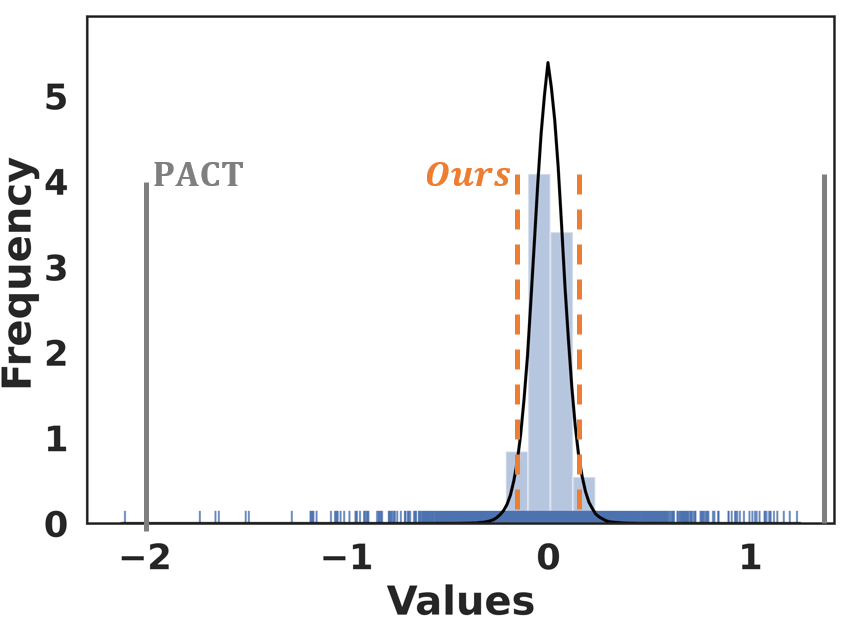}
		\label{fig:wo4}
	}
	\subfigure[$\m w_g$  at Layer 4.]{
		\includegraphics[width=0.22\textwidth]{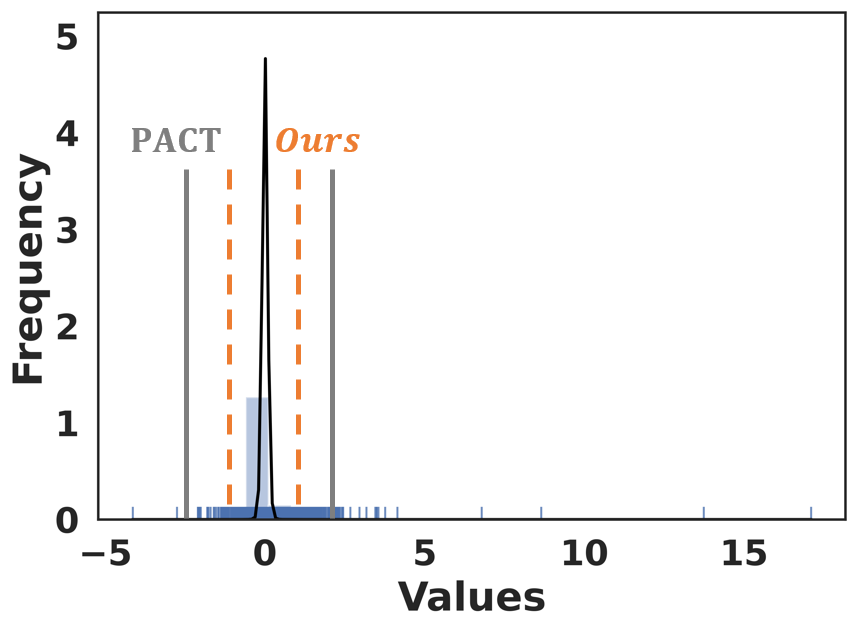}
		\label{fig:wg4}
	}
\caption{Distributions of 
output projection matrix $\m w_o$ in the multi-head attention module and the second linear layer $\m w_g$ in the feed-forward network of the $4$-th layer from the 12-layer full-precision GPT-2 trained on PTB dataset. 
Other modules in other layers exhibit similar patterns. 
Vertical lines indicate the clipping factors learned by PACT and our method.
Black curves show the estimated distribution by kernel density estimation. 
}
	\label{fig:weight_dist}
\end{figure}

Besides the learned word embeddings, we also investigate the distribution of the weights in the full-precision model.
Figure~\ref{fig:weight_dist} shows that the weight distributions  of a 12-layer full-precision GPT-2
are highly skewed with outliers.
This causes difficulty in estimating the clipping factor $\alpha$ of the quantizer by heuristic methods, 
or even by PACT which learns the $\alpha$ through gradient descent.
Specifically, in PACT,
the approximated gradient of $\alpha$ only relies on the 
weights whose absolute values are larger than $\alpha$.
This solution ignores the effect of weights within $[-\alpha, \alpha]$ and depends heavily on the initialization of $\alpha$.
Figure~\ref{fig:weight_dist} shows that an improper initialization together with the inaccurate 
gradient estimation of the clipping factor often make the learned $\alpha$ of PACT too large, and can not provide fine resolution to the majority of weights within the clipping range. 
The quantization error accumulated over time makes this problem more  severe.  In this work, we re-parameterize the clipping factor to make the quantizer adaptive to each module in the Transformer layers, and consider both weights outside and inside the clipping range when estimating the gradient of the clipping factor.

\begin{figure*}[t]
\vspace{-0.2in}
\centering 
\includegraphics[width=0.9\textwidth]{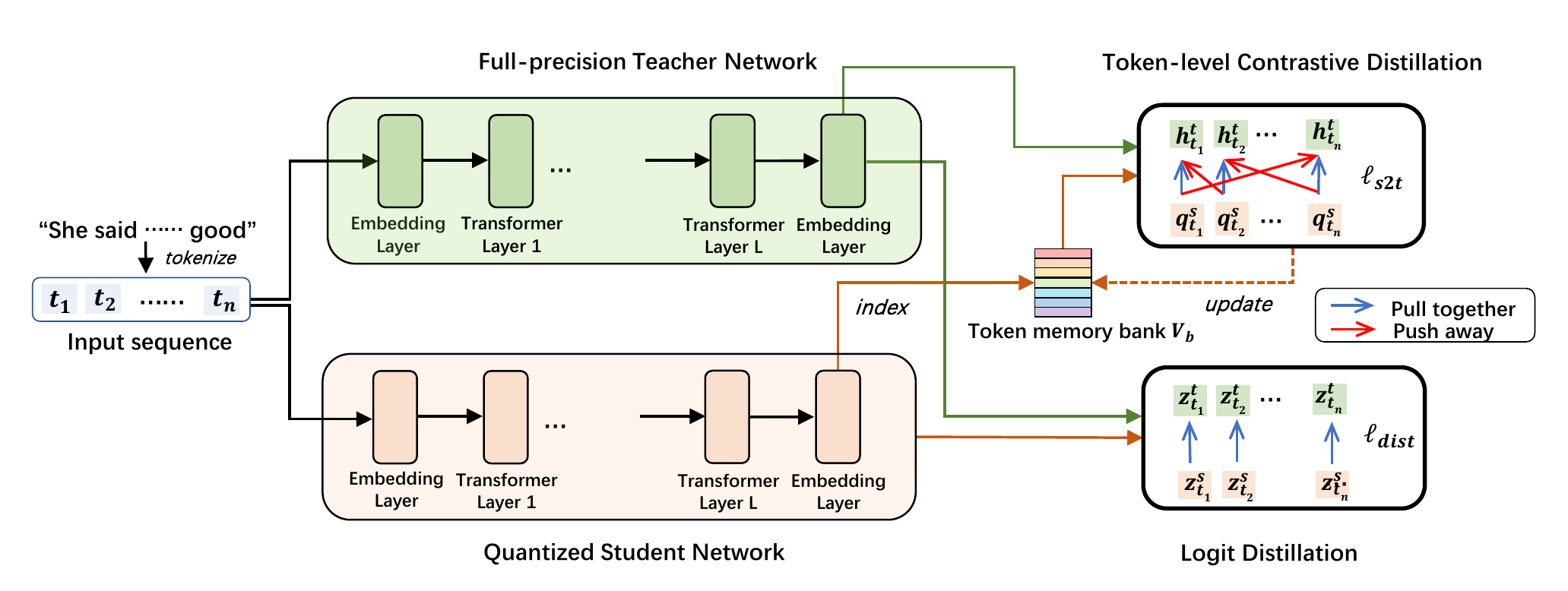}
	\vspace{-0.1in}
\caption{The training workflow of the proposed method. For each token,
we compute both (i) the student-to-teacher and teacher-to-student token-level contrastive distillation loss;
and (ii) the distillation loss on the logits. 
For simplicity of illustration, only the student-to-teacher token-level contrastive loss is shown here.
The embedding layer and all weights in the Transformer layers are quantized with the proposed  module-dependent dynamic scaling.
} 
\label{fig:model}
\end{figure*}

As will be discussed in Section~\ref{sec:dyna_scaling},
we propose a module-wise dynamic scaling to reduce the clipping factor's sensitivity to initialization, and 
an improved gradient estimation that also considers the weights within $[-\alpha, \alpha]$.
Figure~\ref{fig:weight_dist} shows that the clipping factor learned by our  method
gives finer resolutions to the majority of the weights.

\section{Proposed Method}
\label{sec:method}
Based on the observations in Section~\ref{sec:difficulty}, we propose a quantization method which utilizes token-level contrastive distillation to make the word embedding  distinguishable (Section~\ref{sec:seq_contrastive_learning}) and a module-wise dynamic scaling adjustment to learn better clipping factors (Section~\ref{sec:dyna_scaling}).


\subsection{Token-level Contrastive Distillation}
\label{sec:seq_contrastive_learning}
The proposed token-level contrastive distillation 
contrast among tokens instead of sequences
sequence, to learn
distinguishable representations for each token.
Inspired by \citet{baevski2020wav2vec},  which uses in-utterance representation at different positions of the same utterance as negatives for speech feature learning,
for each token of the quantized network, we use the representation of the same token from the  full-precision teacher network as its positive, while representations of other tokens in the same sequence as negatives (Figure~\ref{fig:model}).
Inspired by \citet{he2020momentum}  which uses a momentum encoder for more consistent representation,
we  build a memory bank 
to store momentum token representations.
When computing the contrastive distillation loss, we load the  representations of negative samples from the memory bank with cheap indexing operations.

Specifically, we use
superscripts $s$ and $t$ to denote the quantized student network and full-precision teacher network, respectively.
Denote the length-$n$ input sequence of tokens as $(t_1, t_2, \cdots, t_n)$.
For  the $i$-th token $t_i$, 
suppose its hidden states of the last Transformer layer from  the quantized and full-precision network  are linearly projected to 
$(\h_{t_i}^s, \h_{t_i}^t) \in \mathbb{R}^{d}$,
and  $\q_{t_i}^s$ 
is the smoothed representation of 
$\h_{t_i}^s$
in the memory bank. 
Denote $\mS_i$ as the union set of the index $i$ of the positive and the indices of the sampled negatives for token $i$, 
the student-to-teacher
token-level contrastive loss for the length-n sequence
can be formulated as 
\begin{equation}
    \label{eq:loss_contrastive_loss}
    \!\!\ell_{s2t} \!=\!
    -\sum_{i=1}^{n} \!\log \frac{\exp(s(\q_{t_i}^s, \h_{t_i}^t)/\tau )}{\sum_{j \in \mS_i} \exp(s(\q_{t_i}^s, \h_{t_j}^t)/\tau )},
\end{equation}
where $s(\x,\y) = \frac{\x^\top\y}{\|\x\|\|\y\|}$ computes  the cosine similarity, and $\tau$ is a fixed temperature parameter.
Similarly, we also 
use a memory bank to store momentum token representations for the teacher network, and compute a teacher-to-student contrastive loss 
$\ell_{t2s}$.
The final contrastive loss is the average of the contrastive losses from both sides, i.e., 
$\ell_{cont} = \frac{1}{2} (\ell_{s2t} + \ell_{t2s})$.
We update the representation of token $t_i$ in the  memory bank with the moving-average of token representations from the quantized network:
\begin{equation}
\q_{t_i}^s \leftarrow m\q_{t_i}^s + (1-m) \h_{t_i}^s,
\label{eq:mem_bank}
\end{equation}
where $m \in [0,1)$ is the momentum coefficient that controls the smoothness of the token representation. The representation of teacher used in $\ell_{t2s}$ is also a smoothed version of $\h_{t_i}^t$ via moving-average with  the momentum coefficient $m$.

Besides,
we use an additional  distillation loss 
$\ell_{dist}$
over the logits.
For  the $i$-th token $t_i$, suppose the logits  of the quantized and full-precision network are $\z_{t_i}^s, \z_{t_i}^t\in \mathbb{R}^{|V|}$.
$\ell_{dist}$
is computed with the soft cross-entropy loss:
\begin{equation}
    \label{eq:distillation_loss}
    \ell_{dist}
    = - \sum_{i=1}^{n} \z_{t_i}^t \log(\z_{t_i}^s).
\end{equation}
Thus the total training loss is 
\begin{equation}
    \label{eq:loss}
    \ell = \lambda \ell_{cont} + \ell_{dist},
\end{equation}
where $\lambda$ is a trade-off factor set as 0.1 by default. 

Intuitively, for each token in the quantized network, $\mL_{dist}$ only encourages it to mimic its corresponding token of the teacher network, while
$\mL_{cont}$ 
not only pulls it close to its positive, but also pushes it away from its negatives.
In this way, $\mL_{cont}$  helps the student to capture more information from the teacher’s representation, as is also theoretically discussed in \citet{tian2019contrastive}.

The proposed token-level contrastive distillation outperforms the sequence-level counterpart (as will be shown empirically in Section \ref{section-neg}).
We conjecture this is because (i) token-level contrast alleviates the problem of homogeneous word embedding (Figure~\ref{fig:embedding_collapse})
in the low-bit quantization; and (ii) similar to speech, the order of natural language is also sequential instead of spatial like images; and (iii) the self-attention mechanism allows other tokens to learn representations contextualized on the studied token,  and these in-sequence negatives are harder than those from in-batch sequences,  allowing more efficient representation learning.




\subsection{Module-dependent Dynamic Scaling}
\label{sec:dyna_scaling}



Based on the observation of varied  weight distribution in Section~\ref{sec:difficulty},  we propose a simple-yet-effective dynamic scaling
according to the statistics of each module weight. 
Specifically,  instead of directly learning the original clipping factor $\alpha$ as PACT,
we turn to learn a new scaling factor  $\gamma$, which is multiplied with the average weight magnitude $\frac{\|\m w\|_1}{n}$ to get clipping factor $\alpha$:
\begin{equation}
    \label{eq:new_scaling}
    \alpha = \gamma  \cdot \frac{\|\m w\|_1}{n},
\end{equation}
where $\|\cdot\|_1$ denotes $\ell_1$ norm.
With a slight abuse of notation, here we use $n$ to denote the number of elements in $\m w$.
The scaling $\gamma$ is initialized as 1, which not only eases the initialization but also ensures the initial clipping factor $\alpha$ does not deviate far from the full-precision weights,
regardless of the diversity of weight distribution. 

Besides, we design a more accurate gradient estimation of the scaling factor than PACT \citep{choi2018pact}.
Previous PACT only back propagates through weights whose absolute values are 
larger than or euqal the clipping factor (\textit{i.e.}  $w \geq \alpha$).
Instead, we also consider the weights inside the clipping range  
(\textit{i.e.} 
$ w < \alpha$).
In particular, the gradient w.r.t $\gamma$ is the summation of contributions from each weight, i.e., $\frac{\partial \ell}{\partial \gamma} = \sum_{i=1}^n \frac{\partial \ell}{\partial [\m w_q]_i}\frac{\partial [\m w_q]_i}{\partial \gamma}$.
We compute the gradient contributed by each weight $\frac{\partial \ell}{\partial [\m w_q]_i}\frac{\partial [\m w_q]_i}{\partial \gamma}$ as:
\begin{equation}
\begin{aligned}
 \left\{\begin{matrix}
 &\!\!\!\! \frac{\partial \ell}{\partial [\m w_q]_i}Q(u_i) \frac{\|\m w\|_1}{n}, w_i\leq -\alpha\\ 
 &\!\!\!\! \frac{\partial \ell}{\partial [\m w_q]_i}[- \frac{w_i}{\alpha} + Q(u_i) ] \frac{\|\m w\|_1}{n}, -\alpha\!<\! w_i \!<\!\alpha \label{eq:gradient-scaling} \\ 
 &\!\!\!\! \frac{\partial \ell}{\partial [\m w_q]_i}Q(u_i) \frac{\|\m w\|_1}{n}, w_i\geq \alpha 
\end{matrix}\right.
 \end{aligned}
\end{equation}
where $\ell$ is the total training loss and 
$u_i = \text{clip}(w_i, -\alpha, \alpha) / \alpha$. The detailed derivation can be found in Appendix \ref{apdx:scaling_grad}. 

Intuitively,   the update of  clipping factor should be influenced by both weights outside and inside $[-\alpha, \alpha]$,
since $\alpha$ controls the quantization error of both, \textit{i.e.}, a large clipping factor results in small quantization error for weights outside $[-\alpha, \alpha]$, while large error for weights inside.
Our new estimation of the gradient of $\gamma$ 
considers  weights both
outside and inside $[-\alpha, \alpha]$. Additionally, the proposed scaling is less sensitive to the varied distribution of weight than PACT, since the gradient of scaling $\frac{\partial \ell}{\partial  \gamma}$ is proportional to the  average weight magnitude  $\frac{\|\m w\|_1}{n}$.


\section{Experiments}

\begin{table*}[htbp]
\vspace{-0.2in}
    \centering
    \scalebox{0.85}{
    \setlength{\tabcolsep}{5mm}{
    \begin{tabular}{c|cc|ccc|c}
    %
    Method & 	\tabincell{c}{\#Bits\\(W-E-A)} & \tabincell{c}{Size\\ (MB) ($\downarrow$)} &  \tabincell{c}{WikiText2\\PPL ($\downarrow$)} & \tabincell{c}{PTB\\PPL ($\downarrow$)} & \tabincell{c}{WikiText103\\PPL ($\downarrow$)} & \tabincell{c}{Persona-Chat\\Acc(\%)  ($\uparrow$) } \\
    \toprule
    - & \textit{full-prec.} & 474.9 & 14.48&		14.72	&	14.19 & 77.01\\
    \hline\hline
    \textit{PACT} & 8-8-8 &  121.4 & 17.49	&	16.11	&	16.76 & 74.73\\
    \textit{LSQ} & 8-8-8 &   121.4 & 16.75	&	15.43	&	15.24 & 75.28\\
    \textit{LAQ} & 8-8-8 &   121.4 & 16.91	&	15.87&		15.88& 76.02\\
   \hdashline
    \textit{QuantGPT}  & 8-8-8 & 121.4& \textbf{15.31}	&\textbf{14.90}&		\textbf{14.58} & \textbf{76.12}\\

    \hline
    \textit{PACT} & 4-4-8 & 62.4 &19.23	&	20.17&		20.15 & 25.13\\
    \textit{LSQ} & 4-4-8 &62.4  & 78.99	&	79.76	&	75.12& 45.10\\
    \textit{LAQ}& 4-4-8 &62.4  & 17.12	&	16.55	&	16.91& 71.71\\
    \hdashline
    \textit{QuantGPT} & 4-4-8 & 62.4 & \textbf{15.55}	&	\textbf{14.95}&		\textbf{15.31}&\textbf{76.57}\\

    \hline
    \textit{PACT} & 2-2-8 &  33.0 & 173.02	&	189.13	&	171.03 & 5.52\\
    \textit{LSQ} & 2-2-8 & 33.0 & 847.54&		544.98	&	1470.86& 5.54\\
    \textit{LAQ}& 2-2-8 & 33.0 &19.15	&	18.25&		18.97 & 71.36\\
    \hdashline
    \textit{QuantGPT} & 2-2-8 & 33.0 &\textbf{17.30}		&\textbf{16.12}	&	\textbf{16.98}&  \textbf{74.78}\\
    \end{tabular}}}
    \vspace{-0.05in}
    \caption{Results of language modeling  on the test set of WikiText2, PTB and WikiText103 datasets, 
    and next utterance prediction on the validation set of Persona-Chat dataset,
    with quantized GPT-2.
``\#Bits (W-E-A)'' represents the bit-width for weights of Transformer
layers, word embedding, and activations. 
    }
    \label{tbl:gpt2_lm}
    \vspace{-0.1in}
\end{table*}

\label{sec:expt}

\subsection{Setup}
\paragraph{Tasks and Models.}
In this section, we evaluate the efficacy of our proposed quantization method on three 
kinds of generative 
tasks on two kinds of generative pre-training models.
Specifically, we  perform the proposed quantization approach  on language modeling and next utterance prediction tasks on GPT-2 \citep{gpt}, and  abstractive summarization using 
BART~\citep{bart}, and call the resultant  models  QuantGPT and QuantBART.
The token-level contrastive distillation is performed on the hidden states of the last layer of GPT-2 or the BART decoder.
More details about the datasets and  model architectures can be found in Appendix~\ref{apdx:dataset} and ~\ref{apdx:model_arch}.

\paragraph{Implementation Details.}
For each downstream task with our proposed method, 
we first fine-tune a full-precision network using the pre-trained checkpoint from huggingface\footnote{\url{http://huggingface.co/models}} 
for both GPT-2 and BART.
Then we use this fine-tuned network as the full-precision teacher network and to initialize the quantized student network.
We train each task with 8 V100 GPUs based on the Pytorch framework. 
The detailed hyper-parameters for each task are available in Appendix~\ref{apdx:hyperparam}.


\paragraph{Compared Methods.}
Since there are very few attempts to compress generative PLMs,
we  self-implement three 
baseline quantization methods
PACT~\citep{choi2018pact}, LSQ \citep{esser2020learned} and LAQ \citep{hou2018loss} for comparison.
Details about these  methods are in Appendix \ref{apdx:q-method}. 

\subsection{Language Modeling}

\setlength{\tabcolsep}{2.5mm}

The task of language modeling is to predict the probability distribution over a sequence of words.
For language modeling, we experiment on WikiText2 \citep{merity2016pointer}, Penn Treebank (PTB) \citep{mikolov2012context} and WikiText103 \citep{merity2016pointer}.
We use perplexity (PPL) to evaluate the performance for language modeling.  

\paragraph{Comparison with the Full-precision Model.}
From Table \ref{tbl:gpt2_lm}, 
the performance of the proposed method  with 8-bit weight is comparable to the full-precision counterpart on PTB and WikiText103, while drops slightly on WikiText2.
A slightly more severe performance drop is observed as the bit-width decreases from 8 to 4, with a drop of around 1 PPL point on WikiText2 and WikiText103, and less than 0.1 PPL point on PTB.
When the bit-width  of weight further goes down to 2, our method has an average of 2 PPL points drop, but
achieves 14.4$\times$ model size reduction. 

\paragraph{Comparison with Other Quantization Methods.}
From Table~\ref{tbl:gpt2_lm}, our method outperforms PACT, LSQ and LAQ
for all bit-widths and tasks.  
As the bit-width decreases from 8 to 4,  
the PPL of LSQ greatly increases, with the average PPL of LSQ increasing by over 5 times.
As the bit-width further decreases to 2, both LSQ and PACT fail on all datasets, despite their good performance on understanding tasks on BERT~\citep{binarybert}.
We conjecture it is because though 
both PACT and LSQ have learnable parameters,
the accumulated quantization error of generative PLMs makes the updates of these parameters by gradient descent less stable. 
On the other hand,  
the proposed module-wise dynamic scaling alleviates the problem.
\begin{table}[htbp]
    \centering
    	\scalebox{0.85}{
    	   \setlength{\tabcolsep}{0.43mm}{
    \begin{tabular}{c|c|ccc}
     Method & \tabincell{c}{Size\\(MB)($\downarrow$)} &  \tabincell{c}{WikiText2\\PPL($\downarrow$)} & \tabincell{c}{PTB\\PPL($\downarrow$)} & \tabincell{c}{WikiText103\\PPL($\downarrow$)}  \\
    \toprule
     \textit{full-prec.} & 474.9 (1.0x) & 14.5 & 14.7 &14.2 \\
    \hline\hline 
     KnGPT2 & 332.0 (1.4x)&  -&- &  20.5 \\
    \hline
    DistilGPT2  &329.6 (1.4x)&-&- &  21.1\\
    LightPAFF  &268.0 (1.8x)&18.8 &22.8  &16.4 \\
    \hline
    \textit{Ours(8-8-8)} &121.4 (3.9x)& \textbf{15.3}	&	\textbf{14.9}	&	\textbf{14.6}\\
    \textit{Ours(4-4-8)} &62.4 (7.6x)& 15.6	&	15.0	&	15.3\\
    \textit{Ours(2-2-8)} &33.0 (\textbf{14.4x})& 17.3	&	16.1	&	17.0\\
    \end{tabular}}}
    \vspace{-0.05in}
    \caption{Comparison between our proposed quatization method 
    and other compression methods on GPT-2. 
    }
    \vspace{-0.8mm}
    \label{table-compare-others}
    \vspace{-0.1in}
\end{table}

\paragraph{Comparison with Other Compression Methods.}
In  Table \ref{table-compare-others}, we compare our quantization method against
recent  GPT-2 compression methods,
including tensor decomposition method KnGPT2 \citep{edalati2021kronecker}, as well as distillation methods DistilGPT2 and LightPAFF~\citep{song2020lightpaff}. 
From the comparison, our method outperforms the others in terms of model size and performance, even when weights are compressed to only 2 bits.

\subsection{Next Utterance Prediction}
The task of next utterance prediction predicts the next utterance given the dialogue context. 
It tests the language understanding ability of generative models. 
For this task, we use a large-scale dialogue dataset, Persona-Chat \citep{zhang2018personalizing}.
From Table \ref{tbl:gpt2_lm}, all
quantization methods incur a clear performance  drop  
compared to the full-precision baseline,
even in the 8-bit setting.  
As the quantization becomes more aggressive, \textit{i.e.}, the bit-width gets smaller, 
the performance of PACT and LAQ decrease more significantly than ours. 
In particular, LSQ diverges for 2-bit weight and 
its accuracy is only 5\%, which is no better than a random guess
as there are 20 classes.

\subsection{Abstractive Summarization}
\label{sec:abs_summarization}
Abstractive summarization aims at generating a terse summary that captures the main ideas of the source article. 
We experiment on XSum \citep{narayan2018don}, 
whose ground-truth summarizations are  highly abstractive
and are challenging for many 
extractive strategies.
ROUGE 1, 2, L are used to evaluate the performance of this task.

\begin{table}[htbp]
    \centering
	\scalebox{0.85}{
	    \setlength{\tabcolsep}{0.95mm}{
	\begin{tabular}{c|cc|cccc}
    Method & \tabincell{c}{\#Bits\\(W-E-A)} & \tabincell{c}{Size\\(MB)($\downarrow$)} &  \multicolumn{3}{c}{XSum}   \\
    \toprule 
    Metric & &  &R1 ($\uparrow$)&R2 ($\uparrow$) &RL ($\uparrow$)   \\
    \hline
    - & \textit{full-prec.} &    532.0  &	40.75&	18.10&	33.05 \\
    \hline\hline
    \textit{PACT} & 8-8-8 &  138.1   &	39.16&	16.60&	31.60 \\
    \textit{LSQ} & 8-8-8 & 138.1     &	39.09&	16.72&	31.56\\
    \textit{LAQ}& 8-8-8 &   138.1  &	39.10&	16.74&	31.65 \\
    \hdashline
    \textit{QuantBART}& 8-8-8&   138.1  &	\textbf{40.25}&	\textbf{17.78}&	\textbf{32.70} \\

    \hline
    \textit{PACT} & 4-4-8  &72.4	&32.68	&11.52	&26.03\\
    \textit{LSQ}  & 4-4-8   & 72.4&	38.94&	16.48&	31.46\\
    \textit{LAQ} & 4-4-8    &72.4 &39.03&	16.68&	31.63 \\
    \hdashline
    \textit{QuantBART} & 4-4-8   & 72.4 &	\textbf{40.24}&	\textbf{17.71}&	\textbf{32.69}\\

    \hline
    \textit{PACT} & 2-2-8& 39.6  &	7.76&	1.30	&6.96\\
     \textit{LSQ} & 2-2-8    & 39.6&	37.09&	14.88	&29.76\\
    \textit{LAQ} & 2-2-8&  39.6  &	37.48	&15.27&	30.13\\
    \hdashline
    \textit{QuantBART}  & 2-2-8 & 39.6  &	\textbf{39.15}	&\textbf{16.72}&	\textbf{31.72}\\
    \end{tabular}}}
    	\vspace{-0.05in}
    \caption{Results of abstractive summarization on the test set of the XSum dataset,  with quantized BART.
    }
    \label{table-bart-base}
\end{table}

Table \ref{table-bart-base} shows the results of the abstractive summarization. 
As can be seen, 
our  method constantly 
outperforms other  methods again
with a clear margin.   
Example generated summarizations of different methods 
in Appendix~\ref{apdx:example_summarization} show that the summaries generated by QuantBART are logical and terse, while those from PACT have  repeated texts.

\section{Discussion}
\label{sec:discussion}
\subsection{Ablation on Contrastive Learning}
\subsubsection{Choices of Negative Sampling}
\label{section-neg}

\begin{figure}[htbp]
    \centering
	\subfigure[fp+quan.]{
		\includegraphics[width=0.14\textwidth]{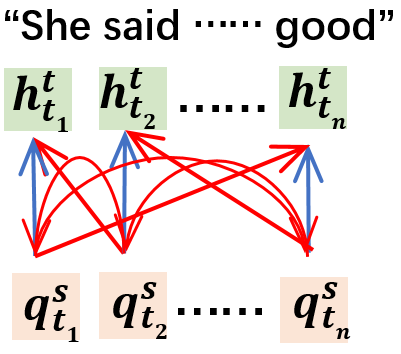}
		\label{fig:neg_mode1}
	}
	\subfigure[quan. only.]{
		\includegraphics[width=0.14\textwidth]{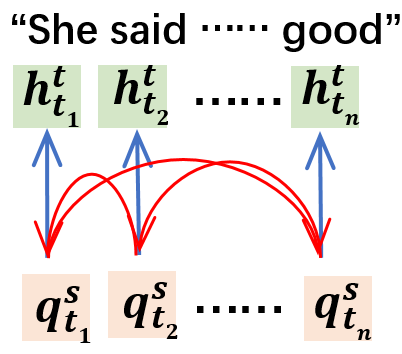}
		\label{fig:neg_mode3}
	}
	\vspace{-0.1in}
	\subfigure[global.]{
		\includegraphics[width=0.28\textwidth]{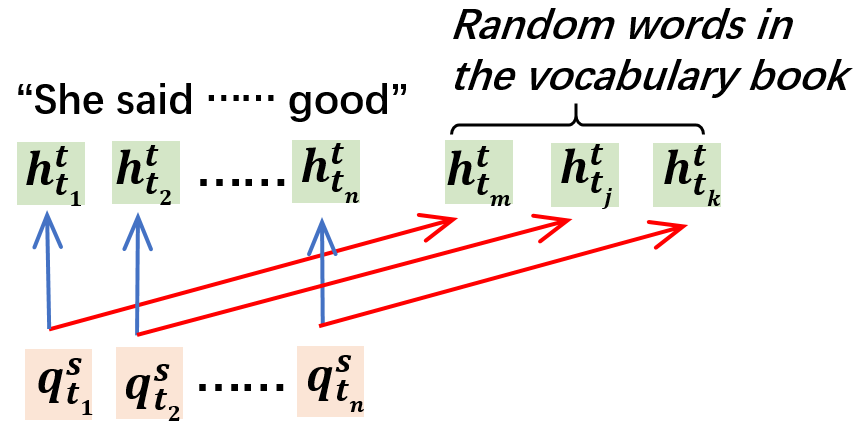}
		\label{fig:neg_global}
	}
	\subfigure[in-batch.]{
		\includegraphics[width=0.14\textwidth]{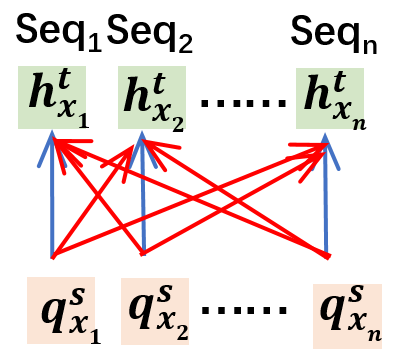}
		\label{fig:neg_seqlevel}
	}
\caption{Four variants of negative sampling, illustrated using the student-to-teacher contrastive loss.
}
\label{fig:ablation_neg}
\end{figure}

As shown in Figure \ref{fig:ablation_neg}, we ablate on how to choose negative samples in contrastive learning.
For simplicity of illustration, we only elaborate on the choices of negatives for the  student-to-teacher contrastive loss $\mL_{s2t}$, and choose negatives for the teacher-to-student contrastive loss $\mL_{t2s}$ in a similar way.
Specifically, 
we compare our method with  
variants of token-level contrastive learning, which select negative samples of each token of the quantized student network from
(a) representations of other tokens in both the full-precision and quantized networks (\textit{fp+quan.}); 
(b) representations of other tokens in the quantized network (\textit{quan. only})
and
(c) the whole vocabulary randomly for each training iteration (\textit{global}).
Besides, we compare with (d) sequence-level contrastive learning by pulling together representations of the same sequence, and pushing away representations of different ones from the teacher network (\textit{in-batch}).  
Representation of a sequence is defined as the mean of representations of all tokens in the sequence. 

From Table \ref{table-ablation-neg}, \textit{``fp+quan.''} and \textit{``quan. only''}
performs worse than QuantGPT, which uses full-precision representations of other tokens as negative samples. 
This indicates that noisy 
representations of tokens from the not-fully-trained quantized network may not be sufficient.
\textit{``global''}  performs even worse, which we conjecture is because, for one token, negative tokens chosen from the same sequence are contextually related to it and more informative than random tokens. 
\textit{``in-batch''} performs worse than all token-level variants, which may be because generative tasks make predictions in a token-wise manner and rely heavily in finer-grained token-wise representations.
Interestingly, contrary to in-batch negative sampling in computer vision~\citep{chen2020simple}, we find that reducing the number of negative samples by reducing the batch size from 32 to 16 slightly improves performance.

\begin{table}
    \centering
    \scalebox{0.85}{
    	    \setlength{\tabcolsep}{0.42mm}{
    \begin{tabular}{c|c|ccc}
    - & \tabincell{c}{Sampling\\method} &WikiText2  & PTB & WikiText103\\
    \toprule
    - & \textit{\textit{QuantGPT}} & \textbf{17.30}	&	\textbf{16.12}	&	\textbf{16.98} \\
    \hline\hline
     \multirow{3}{*}{Tok-level}   & fp+quan.   & 17.38	&	16.51	&	17.13\\
     & quan. only  &17.35	&	16.54	&	17.15\\
     & global  & 17.71	&	16.63	&	17.55\\
     \hline
    \multirow{2}{*}{Seq-level} & in-batch (bz=32) &17.62	&	19.23	&	18.97\\
     & in-batch (bz=16) &17.48	&	17.11	&	18.16\\
    \end{tabular}}}
        \vspace{-0.05in}
    \caption{Ablation study on negative sampling for 2-bit weight, ``bz'' denotes for the batch size. ``Tok'' and ``Seq'' are abbreviation for token and sequence, respectively. }
    \label{table-ablation-neg}
    \vspace{-0.1in}
\end{table}


\begin{figure}[htbp]
    \centering
    \begin{minipage}{0.236\textwidth}
        \centering
		\includegraphics[width=0.95\textwidth]{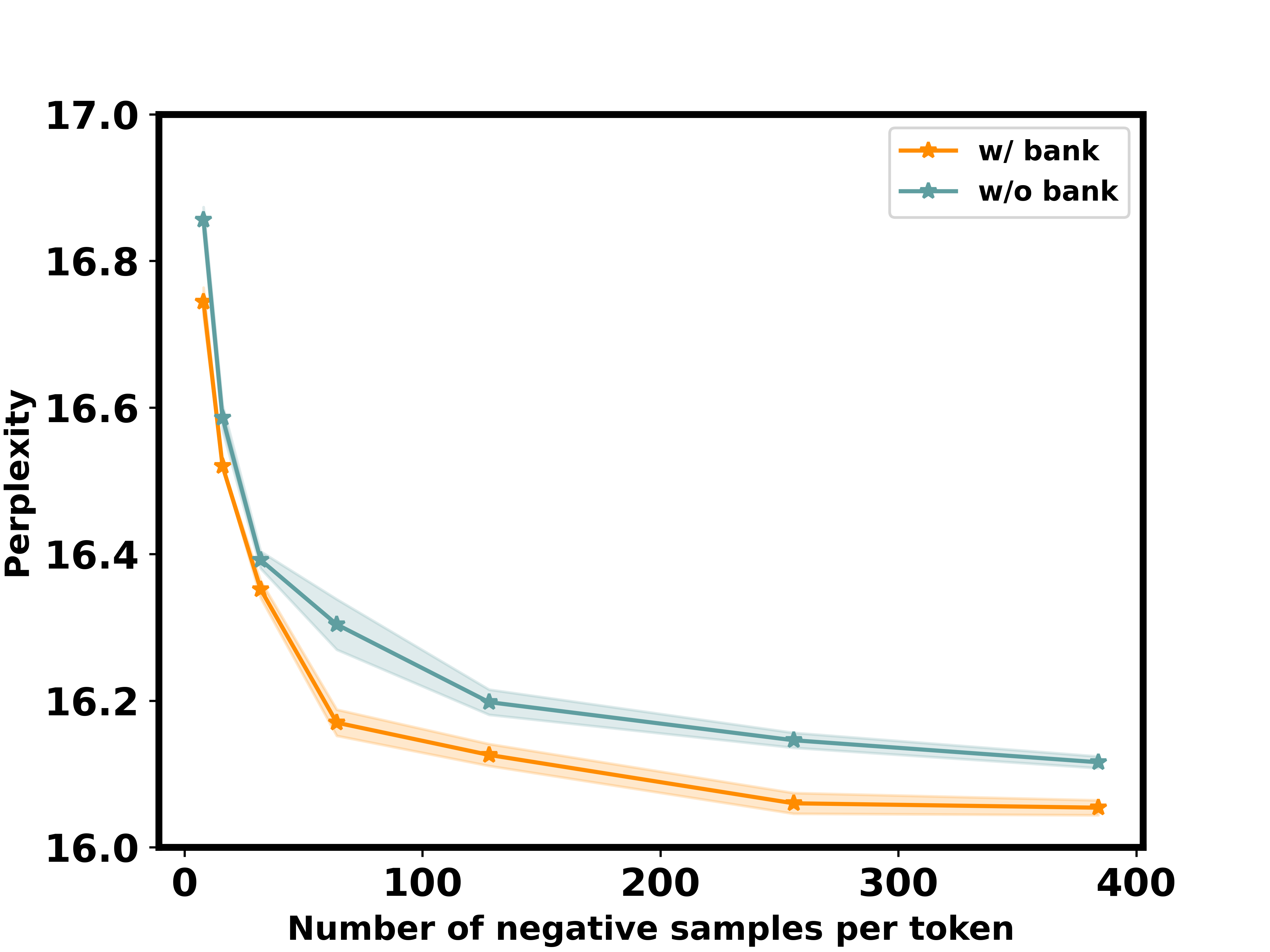}
        \caption{Effect of the number of negative samples.}
        \label{fig:ablation_bank}
    \end{minipage}%
    \hspace{0.05in}
    \begin{minipage}{0.229\textwidth}
        \centering
		\includegraphics[width=0.95\textwidth]{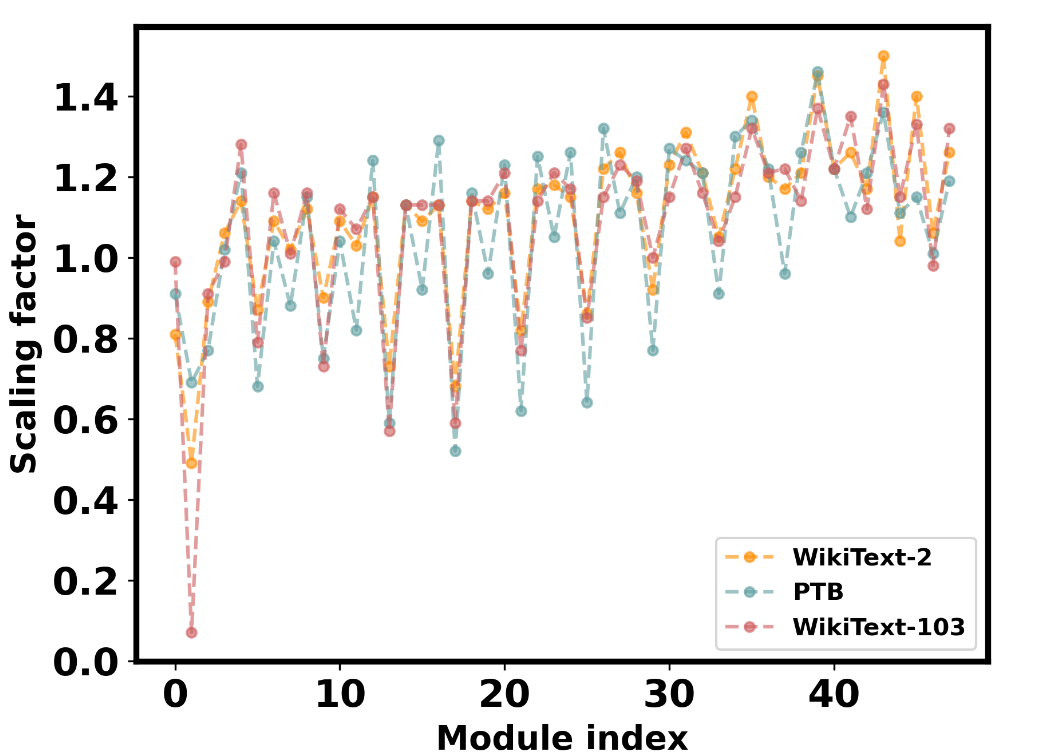}
			\vspace{-0.08in}
        \caption{ Scaling factors in the 2-bit QuantGPT.}
        \label{fig:scaling}
    \end{minipage}
\end{figure}

\subsubsection{Number of Negative Samples}
In Figure \ref{fig:ablation_bank}, we plot the PPL of  2-bit \textit{QuantGPT} on the PTB dataset, with varying number of negative samples. 
We plot the mean results with standard deviations from 5 independent runs.
As can be seen, the performance improves and converges gradually  as the number of negative samples increases. Figure \ref{fig:ablation_bank} also shows that 
using the moving-average representations ($\q^s_{t_i}$ in Eq. \eqref{eq:mem_bank}) of negative samples
in the memory bank has better performance than using the immediate representations  ($\h^s_{t_i}$ in  Eq. \eqref{eq:mem_bank}),
because of a smoother and more consistent representation of tokens.

\subsubsection{Training Cost of the Contrastive Loss}
In Table \ref{table-cost}, we report the training
speed and memory consumption of
training the GPT-2 model on the PTB dataset with and without
the proposed
 token-level contrastive loss.
 Batch size is set to 4 per device, which can be increased by using GPUs with larger memory or reducing the sequence length  of samples.
As can be seen, 
with the proposed token-level contrastive loss, the performance clearly improves with only slightly slower training speed and more memory consumption.

\begin{table}
    \centering
    \scalebox{0.85}{
    	    \setlength{\tabcolsep}{2mm}{
    \begin{tabular}{c|c|c|c}
     \tabincell{c}{ Training loss} &  \tabincell{c}{ Training time\\ (sec/iter) ($\downarrow$)}   & \tabincell{c}{ Memory\\ (MB) ($\downarrow$)} & PPL ($\downarrow$)\\
    \toprule
    $\ell_{dist}$  &  0.61 & 14700&16.93 \\
    \hline
    $\lambda \ell_{cont} + \ell_{dist}$ 
    & 0.67 &14839 & 16.12\\
    \end{tabular}}}
    \caption{Efficiency study of the token-level contrastive learning. The results are reported on the PTB dataset on 2-bit GPT-2. 
    ``sec/iter'' means the needed time in seconds per iteration. Memory denotes the GPU consumption per device.}
    \label{table-cost}
\end{table}

\subsubsection{Representations for the Contrastive Loss}
In Table \ref{table-contrast-pos}, we compare the different representations to perform the contrastive loss. 
The ``decoder-last''( resp. ``decoder-first'')  denotes performing the proposed token-level contrastive loss  on the hidden states from the 
last decoder layer (resp. first decoder layer)
followed by
a linear transformation.

From Table \ref{table-contrast-pos},  ``decoder-last'' performs better than ``decoder-first''.
A possible reason is that the hidden states of the last decoder blocks contain rich information from all previous layers \citep{xiong2020layer}.
Since the experiments of abstractive summarization are conducted on BART, 
which has both encoder and decoder layers, we also study the contrastive loss on the ``encoder-last'' and ``encoder-first''. 
In the ablation on the encoder, the contrastive loss  $\ell_{cont}$
are computed on the source input (articles), instead of target input (summaries). 
From Table  \ref{table-contrast-pos},  ``decoder-last'' also has better ROUGE 1, 2, L values than other counterparts. 
\begin{table*}[thb]
\setlength{\tabcolsep}{2mm}
    \centering
   	\scalebox{0.9}{
	    \setlength{\tabcolsep}{3.5mm}{
    \begin{tabular}{c|ccc|c|cccc}
    - & WikiText2  & PTB & WikiText103 & Persona-Chat  &\multicolumn{3}{c}{XSum}   \\
    \toprule
     Metric& PPL ($\downarrow$) &PPL ($\downarrow$)&PPL ($\downarrow$)&Acc($\%$) ($\uparrow$)&R1 ($\uparrow$)&R2 ($\uparrow$) & RL ($\uparrow$)  \\
    \hline
     decoder-last &\textbf{17.30}	&\textbf{16.12}&	\textbf{16.98}  &\textbf{74.78}&	\textbf{39.15}&	\textbf{16.72}&	\textbf{31.72} \\
    \hline\hline
    decoder-first & 18.02&	16.61&17.25& 74.75 &39.11&	16.70	&31.62 \\
     encoder-last &-&-&-&-&	38.91&	16.72&	31.67\\
      encoder-first &-&-&-&-&	38.87&	16.70&	31.56\\
    \end{tabular}
    }}

    \caption{Representations for the contrastive loss  
    $\ell_{cont}$
    in 2-bit setting. The ``decoder-last'' means the contrastive loss is computed on the hidden states from the last Transformer layer of the decoder after a linear transform. The naming format works for other variants.}
    \label{table-contrast-pos}

\end{table*}

\subsection{Ablation on Dynamic Scaling}
\label{expt:dyna_scaling}
Figure \ref{fig:scaling} shows the learned scaling  $\gamma$ of different modules in the 2-bit GPT-2 model. 
As can be seen, the scalings  of different modules vary a lot, verifying the need for module-wise dynamic scaling. 

In addition, we investigate the effect of the proposed dynamic scaling 
and the new estimation of the gradient in Eq. ~\eqref{eq:gradient-scaling} with two variants:
1) \textit{$\ell_{dist}$ 
only} which removes the token-level contrastive learning; and
2) \textit{Ours with PACT} which removes the contrastive learning, and estimates the gradient with PACT which only considers the weights whose absolute values are larger than the clipping factor $\alpha$.
As shown in Table \ref{table-ablation-scaling}, the performance gets slightly worse without contrastive learning to regularize to learn distinguishable representations of tokens, but
drops significantly when using PACT to estimate the gradient of the proposed scaling, especially for the WikiText103 dataset. This also indicates that
 the final  performance
gain
 of QuantGPT is mainly attributed to the proposed more accurate quantizer with
 the module-wise dynamic scaling.


\begin{table}[htbp]
    \centering
    \scalebox{0.85}{
    \setlength{\tabcolsep}{2.2mm}{
    \begin{tabular}{c|ccc}
   Method &WikiText2  & PTB & WikiText103\\
    \toprule
   \textit{\textit{QuantGPT}} & \textbf{17.30}	& \textbf{16.12}&	\textbf{16.98} \\
    \hline\hline
    $\ell_{dist}$
    only &  17.85	&	16.93	&	17.78 \\
    Ours with PACT & 20.03	&17.78	&25.54 \\
    \end{tabular}}}
    \caption{Ablation study on the learning of the clipping factor with 2-bit GPT-2 on the language modeling task.
    }
    \label{table-ablation-scaling}
\end{table}

\section{Related Work}
\paragraph{Compression of Generative Pre-trained Language Models.}
Some early explorations compress the generative pre-trained language models.  
KnGPT2 \citep{edalati2021kronecker} applies the Kronecker decomposition to compress the GPT.
DistilGPT2 \footnote{\url{https://transformer.huggingface.co/model/distil-gpt2}} distills a 12-layer GPT-2 to a 6-layer one, which 
is twice as fast during inference.
LightPAFF~\citep{song2020lightpaff} proposes a distillation approach 
that the training loss is a combination of a maximum likelihood loss of the student model, and the KL divergence between the output of teacher and student models. 
SpAtten \cite{wang2021spatten} proposes a sparse model with algorithm and architecture co-design, which removes uninformative tokens and attention heads. Compared with these methods, we not only study the difficulties of compression from the properties of generative tasks, but also study both decoder and encoder-decoder generative models.

\paragraph{Quantization of Pre-trained Language Models.}
Quantization compresses a model by representing the 32-bit floating-point parameter  with a low-bit  representation, 
and has been widely used in various domains 
as it does not require designing a new model architecture.
There have been many attempts to quantize task-specific BERT models~\citep{zafrir2019q8bert,shen2019q,zadeh2020gobo} with only negligible performance drop on natural language understanding tasks.
Recent works~\citep{ternarybert,binarybert} even push the weight bit-width down to as low as 1-bit. 
Despite the success of these approaches for BERT models, attempts to quantize generative PLMs are scarce, and the underlying difficulty remains unclear.

\paragraph{Contrastive Learning.}
Contrastive learning aims at pushing the representations of similar samples together while pulling those of dissimilar ones apart. and
is widely used for large-scale self-supervised learning in various domains~\citep{chen2020simple,sun2020contrastive,baevski2020wav2vec, huang2022spiral}, and multi-modal learning
~\citep{radford2021learning,jia2021scaling}.
SimCLR~\citep{chen2020simple} directly uses other in-batch samples
as negatives,
and sufficient large batch size is required to work well.
MoCo~\citep{he2020momentum} 
maintains a large number of negative samples in a queue and uses a moving average key encoder to improve consistency.
Contrastive learning without negative samples is also proposed in BYOL~\citep{grill2020bootstrap} and SimSiam~\citep{chen2021exploring}. 
Contrastive representation distillation \citep{tian2019contrastive} distills the knowledge from the teacher network to the student network by maximizing the mutual information between them.

The closest work with our token-level contrastive distillation is Wav2vec 2.0~\citep{baevski2020wav2vec},
which use in-utterance representations at different positions
as negatives in speech learning.  
Besides the difference in the modality and tasks, our method also
differs from theirs in (1) \textit{Model:} We quantize the model parameters and activations while they do not; (2) \textit{Representation:} For each sample, we use the output of the full-precision and the quantized networks as its two views, while they use the quantized and the contextualized representation.  (3) \textit{Loss:} We calculate loss over all tokens in an auto-regressive manner, while they only calculate
over the masked tokens non-autoregressively. 




\section{Conclusion}
This paper studies
low-bit quantization of generative PLMs.
We find that the difficulty of quantizing generative PLMs lies in  homogeneous word embedding and varied distribution of weights.
To alleviate the two problems, we propose token-level contrastive learning to learn more distinguishable token embeddings, as well as 
a module-dependent dynamic scaling for more accurate quantization.
 Extensive experiments on language modeling, next utterance prediction and abstractive summarization demonstrate the efficacy of our proposed method.  We hope our work sheds a light on the compression of generative PLMs in future exploration.
%


\section{Limitations}
There are several aspects of this work that can be improved in the future.
First,  
the module-dependent dynamic scaling requires the model to store the intermediate variable  $Q(\u)$ to compute the gradient $\frac{\partial \ell}{\partial \gamma}$ 
during the back propagation. This 
slightly increases the memory consumption in training, and reduces the
the batch size used in the experiments. 
Secondly, the robustness and effectiveness of the  token-level contrastive loss can be improved further. The proposed contrastive loss degrades the performance when the negatives are
not properly chosen
or the trade-off factor $\lambda$ in Eq.(\ref{eq:loss})  is large. 
Thirdly, it would be interesting to empirically implement low-bit operators and verify the real speedup of the proposed method in various hardware platforms.

\section*{Acknowledgements}
This work is supported in part by the General Research Fund (GRF) project 17206020, and in part by ACCESS, AI Chip Center for Emerging Smart Systems, Hong Kong SAR.

\bibliography{custom}

\begin{thebibliography}{41}
\expandafter\ifx\csname natexlab\endcsname\relax\def\natexlab#1{#1}\fi

\bibitem[{Baevski et~al.(2020)Baevski, Zhou, Mohamed, and
  Auli}]{baevski2020wav2vec}
Alexei Baevski, Henry Zhou, Abdelrahman Mohamed, and Michael Auli. 2020.
\newblock wav2vec 2.0: A framework for self-supervised learning of speech
  representations.
\newblock In \emph{Advances in Neural Information Processing Systems},
  volume~33.

\bibitem[{Bai et~al.(2021)Bai, Zhang, Hou, Shang, Jin, Jiang, Liu, Lyu, and
  King}]{binarybert}
Haoli Bai, Wei Zhang, Lu~Hou, Lifeng Shang, Jing Jin, Xin Jiang, Qun Liu,
  Michael Lyu, and Irwin King. 2021.
\newblock Binarybert: Pushing the limit of bert quantization.
\newblock In \emph{Annual Meeting of the Association for Computational
  Linguistics}.

\bibitem[{Bengio et~al.(2013)Bengio, L{\'e}onard, and Courville}]{ste}
Yoshua Bengio, Nicholas L{\'e}onard, and Aaron Courville. 2013.
\newblock Estimating or propagating gradients through stochastic neurons for
  conditional computation.
\newblock Technical Report arXiv:1308.3432.

\bibitem[{Brown et~al.(2020)Brown, Mann, Ryder, Subbiah, Kaplan, Dhariwal,
  Neelakantan, Shyam, Sastry, Askell et~al.}]{gpt3}
Tom~B Brown, Benjamin Mann, Nick Ryder, Melanie Subbiah, Jared Kaplan, Prafulla
  Dhariwal, Arvind Neelakantan, Pranav Shyam, Girish Sastry, Amanda Askell,
  et~al. 2020.
\newblock Language models are few-shot learners.
\newblock In \emph{Advances in Neural Information Processing Systems}.

\bibitem[{Chen et~al.(2021)Chen, Tao, and Wong}]{chen2021litegt}
Cong Chen, Chaofan Tao, and Ngai Wong. 2021.
\newblock Litegt: Efficient and lightweight graph transformers.
\newblock In \emph{Proceedings of the 30th ACM International Conference on
  Information \& Knowledge Management}, pages 161--170.

\bibitem[{Chen et~al.(2020)Chen, Kornblith, Norouzi, and
  Hinton}]{chen2020simple}
Ting Chen, Simon Kornblith, Mohammad Norouzi, and Geoffrey Hinton. 2020.
\newblock A simple framework for contrastive learning of visual
  representations.
\newblock In \emph{International conference on machine learning}, pages
  1597--1607.

\bibitem[{Chen and He(2021)}]{chen2021exploring}
Xinlei Chen and Kaiming He. 2021.
\newblock Exploring simple siamese representation learning.
\newblock In \emph{IEEE/CVF Conference on Computer Vision and Pattern
  Recognition}, pages 15750--15758.

\bibitem[{Choi et~al.(2018)Choi, Wang, Venkataramani, Chuang, Srinivasan, and
  Gopalakrishnan}]{choi2018pact}
Jungwook Choi, Zhuo Wang, Swagath Venkataramani, Pierce I-Jen Chuang,
  Vijayalakshmi Srinivasan, and Kailash Gopalakrishnan. 2018.
\newblock Pact: Parameterized clipping activation for quantized neural
  networks.
\newblock Preprint arXiv:1805.06085.

\bibitem[{Courbariaux et~al.(2015)Courbariaux, Bengio, and
  David}]{binaryconnect}
Matthieu Courbariaux, Yoshua Bengio, and Jean-Pierre David. 2015.
\newblock Binaryconnect: Training deep neural networks with binary weights
  during propagations.
\newblock In \emph{Advances in neural information processing systems}, pages
  3123--3131.

\bibitem[{Edalati et~al.(2021)Edalati, Tahaei, Rashid, Nia, Clark, and
  Rezagholizadeh}]{edalati2021kronecker}
Ali Edalati, Marzieh Tahaei, Ahmad Rashid, Vahid~Partovi Nia, James~J Clark,
  and Mehdi Rezagholizadeh. 2021.
\newblock Kronecker decomposition for gpt compression.
\newblock In \emph{Advances in Neural Information Processing Systems}.

\bibitem[{Esser et~al.(2020)Esser, McKinstry, Bablani, Appuswamy, and
  Modha}]{esser2020learned}
Steven~K Esser, Jeffrey~L McKinstry, Deepika Bablani, Rathinakumar Appuswamy,
  and Dharmendra~S. Modha. 2020.
\newblock Learned step size quantization.
\newblock In \emph{International Conference on Learning Representations}.

\bibitem[{Grill et~al.(2020)Grill, Strub, Altch{\'e}, Tallec, Richemond,
  Buchatskaya, Doersch, Pires, Guo, Azar et~al.}]{grill2020bootstrap}
Jean-Bastien Grill, Florian Strub, Florent Altch{\'e}, Corentin Tallec,
  Pierre~H Richemond, Elena Buchatskaya, Carl Doersch, Bernardo~Avila Pires,
  Zhaohan~Daniel Guo, Mohammad~Gheshlaghi Azar, et~al. 2020.
\newblock Bootstrap your own latent: A new approach to self-supervised
  learning.
\newblock In \emph{Neural Information Processing Systems}.

\bibitem[{He et~al.(2020)He, Fan, Wu, Xie, and Girshick}]{he2020momentum}
Kaiming He, Haoqi Fan, Yuxin Wu, Saining Xie, and Ross Girshick. 2020.
\newblock Momentum contrast for unsupervised visual representation learning.
\newblock In \emph{IEEE/CVF Conference on Computer Vision and Pattern
  Recognition}, pages 9729--9738.

\bibitem[{Hendrycks and Gimpel(2016)}]{hendrycks2016gaussian}
Dan Hendrycks and Kevin Gimpel. 2016.
\newblock Gaussian error linear units (gelus).
\newblock Technical Report arXiv:1606.08415.

\bibitem[{Hinton et~al.(2015)Hinton, Vinyals, and Dean}]{hinton2015distilling}
Geoffrey Hinton, Oriol Vinyals, and Jeff Dean. 2015.
\newblock Distilling the knowledge in a neural network.
\newblock Technical Report arXiv:1503.02531.

\bibitem[{Hou et~al.(2020)Hou, Huang, Shang, Jiang, Chen, and Liu}]{dynabert}
Lu~Hou, Zhiqi Huang, Lifeng Shang, Xin Jiang, Xiao Chen, and Qun Liu. 2020.
\newblock Dynabert: Dynamic bert with adaptive width and depth.
\newblock In \emph{Advances in Neural Information Processing Systems},
  volume~33.

\bibitem[{Hou and Kwok(2018)}]{hou2018loss}
Lu~Hou and James~T. Kwok. 2018.
\newblock Loss-aware weight quantization of deep networks.
\newblock In \emph{International Conference on Learning Representations}.

\bibitem[{Hou et~al.(2017)Hou, Quanming, and Kwok}]{hou2017loss}
Lu~Hou, Yao Quanming, and James~T. Kwok. 2017.
\newblock Loss-aware binarization of deep networks.
\newblock In \emph{International Conference on Learning Representations}.

\bibitem[{Huang et~al.(2022)Huang, Zhang, Yeung, Jiang, and
  Liu}]{huang2022spiral}
Wenyong Huang, Zhenhe Zhang, Yu~Ting Yeung, Xin Jiang, and Qun Liu. 2022.
\newblock Spiral: Self-supervised perturbation-invariant representation
  learning for speech pre-training.
\newblock In \emph{International Conference on Learning Representations}.

\bibitem[{Jia et~al.(2021)Jia, Yang, Xia, Chen, Parekh, Pham, Le, Sung, Li, and
  Duerig}]{jia2021scaling}
Chao Jia, Yinfei Yang, Ye~Xia, Yi-Ting Chen, Zarana Parekh, Hieu Pham, Quoc~V
  Le, Yunhsuan Sung, Zhen Li, and Tom Duerig. 2021.
\newblock Scaling up visual and vision-language representation learning with
  noisy text supervision.
\newblock In \emph{International conference on machine learning}.

\bibitem[{Jiao et~al.(2020)Jiao, Yin, Shang, Jiang, Chen, Li, Wang, and
  Liu}]{jiao2020tinybert}
Xiaoqi Jiao, Yichun Yin, Lifeng Shang, Xin Jiang, Xiao Chen, Linlin Li, Fang
  Wang, and Qun Liu. 2020.
\newblock Tinybert: Distilling bert for natural language understanding.
\newblock In \emph{Findings of the Association for Computational Linguistics:
  EMNLP 2020}, pages 4163--4174.

\bibitem[{Lan et~al.(2019)Lan, Chen, Goodman, Gimpel, Sharma, and
  Soricut}]{albert}
Zhenzhong Lan, Mingda Chen, Sebastian Goodman, Kevin Gimpel, Piyush Sharma, and
  Radu Soricut. 2019.
\newblock Albert: A lite bert for self-supervised learning of language
  representations.
\newblock In \emph{International Conference on Learning Representations}.

\bibitem[{Lewis et~al.(2020)Lewis, Liu, Goyal, Ghazvininejad, Mohamed, Levy,
  Stoyanov, and Zettlemoyer}]{bart}
Mike Lewis, Yinhan Liu, Naman Goyal, Marjan Ghazvininejad, Abdelrahman Mohamed,
  Omer Levy, Veselin Stoyanov, and Luke Zettlemoyer. 2020.
\newblock {BART}: Denoising sequence-to-sequence pre-training for natural
  language generation, translation, and comprehension.
\newblock In \emph{Annual Meeting of the Association for Computational
  Linguistics}.

\bibitem[{Loshchilov and Hutter(2017)}]{loshchilov2017decoupled}
Ilya Loshchilov and Frank Hutter. 2017.
\newblock Decoupled weight decay regularization.
\newblock \emph{arXiv preprint arXiv:1711.05101}.

\bibitem[{Merity et~al.(2016)Merity, Xiong, Bradbury, and
  Socher}]{merity2016pointer}
Stephen Merity, Caiming Xiong, James Bradbury, and Richard Socher. 2016.
\newblock Pointer sentinel mixture models.
\newblock \emph{arXiv preprint arXiv:1609.07843}.

\bibitem[{Mikolov and Zweig(2012)}]{mikolov2012context}
Tomas Mikolov and Geoffrey Zweig. 2012.
\newblock Context dependent recurrent neural network language model.
\newblock In \emph{IEEE Spoken Language Technology Workshop}, pages 234--239.
  IEEE.

\bibitem[{Narayan et~al.(2018)Narayan, Cohen, and Lapata}]{narayan2018don}
Shashi Narayan, Shay~B Cohen, and Mirella Lapata. 2018.
\newblock Don’t give me the details, just the summary! topic-aware
  convolutional neural networks for extreme summarization.
\newblock In \emph{Conference on Empirical Methods in Natural Language
  Processing}, pages 1797--1807.

\bibitem[{Radford et~al.(2021)Radford, Kim, Hallacy, Ramesh, Goh, Agarwal,
  Sastry, Askell, Mishkin, Clark, Krueger, and Sutskever}]{radford2021learning}
Alec Radford, Jong~Wook Kim, Chris Hallacy, Aditya Ramesh, Gabriel Goh,
  Sandhini Agarwal, Girish Sastry, Amanda Askell, Pamela Mishkin, Jack Clark,
  Gretchen Krueger, and Ilya Sutskever. 2021.
\newblock Learning transferable visual models from natural language
  supervision.
\newblock In \emph{International Conference on Machine Learning}.

\bibitem[{Radford and Narasimhan(2018)}]{gpt}
Alec Radford and Karthik Narasimhan. 2018.
\newblock Improving language understanding by generative pre-training.

\bibitem[{Raffel et~al.(2020)Raffel, Shazeer, Roberts, Lee, Narang, Matena,
  Zhou, Li, and Liu}]{t5}
Colin Raffel, Noam Shazeer, Adam Roberts, Katherine Lee, Sharan Narang, Michael
  Matena, Yanqi Zhou, Wei Li, and Peter~J Liu. 2020.
\newblock Exploring the limits of transfer learning with a unified text-to-text
  transformer.
\newblock \emph{Journal of Machine Learning Research}, 21:1--67.

\bibitem[{Shen et~al.(2020)Shen, Dong, Ye, Ma, Yao, Gholami, Mahoney, and
  Keutzer}]{shen2019q}
Sheng Shen, Zhen Dong, Jiayu Ye, Linjian Ma, Zhewei Yao, Amir Gholami,
  Michael~W Mahoney, and Kurt Keutzer. 2020.
\newblock Q-bert: Hessian based ultra low precision quantization of bert.
\newblock In \emph{AAAI Conference on Artificial Intelligence}.

\bibitem[{Song et~al.(2020)Song, Sun, Tan, Qin, Lu, Liu, and
  Liu}]{song2020lightpaff}
Kaitao Song, Hao Sun, Xu~Tan, Tao Qin, Jianfeng Lu, Hongzhi Liu, and Tie-Yan
  Liu. 2020.
\newblock Lightpaff: A two-stage distillation framework for pre-training and
  fine-tuning.
\newblock Preprint arXiv:2004.12817.

\bibitem[{Sun et~al.(2020{\natexlab{a}})Sun, Gan, Fang, Cheng, Wang, and
  Liu}]{sun2020contrastive}
Siqi Sun, Zhe Gan, Yuwei Fang, Yu~Cheng, Shuohang Wang, and Jingjing Liu.
  2020{\natexlab{a}}.
\newblock Contrastive distillation on intermediate representations for language
  model compression.
\newblock In \emph{Conference on Empirical Methods in Natural Language
  Processing}, pages 498--508.

\bibitem[{Sun et~al.(2020{\natexlab{b}})Sun, Yu, Song, Liu, Yang, and
  Zhou}]{mobilebert}
Zhiqing Sun, Hongkun Yu, Xiaodan Song, Renjie Liu, Yiming Yang, and Denny Zhou.
  2020{\natexlab{b}}.
\newblock Mobilebert: a compact task-agnostic bert for resource-limited
  devices.
\newblock In \emph{Annual Meeting of the Association for Computational
  Linguistics}, pages 2158--2170.

\bibitem[{Tian et~al.(2019)Tian, Krishnan, and Isola}]{tian2019contrastive}
Yonglong Tian, Dilip Krishnan, and Phillip Isola. 2019.
\newblock Contrastive representation distillation.
\newblock In \emph{International Conference on Learning Representations}.

\bibitem[{Wang et~al.(2021)Wang, Zhang, and Han}]{wang2021spatten}
Hanrui Wang, Zhekai Zhang, and Song Han. 2021.
\newblock Spatten: Efficient sparse attention architecture with cascade token
  and head pruning.
\newblock In \emph{2021 IEEE International Symposium on High-Performance
  Computer Architecture (HPCA)}, pages 97--110. IEEE.

\bibitem[{Xiong et~al.(2020)Xiong, Yang, He, Zheng, Zheng, Xing, Zhang, Lan,
  Wang, and Liu}]{xiong2020layer}
Ruibin Xiong, Yunchang Yang, Di~He, Kai Zheng, Shuxin Zheng, Chen Xing,
  Huishuai Zhang, Yanyan Lan, Liwei Wang, and Tieyan Liu. 2020.
\newblock On layer normalization in the transformer architecture.
\newblock In \emph{International Conference on Machine Learning}, pages
  10524--10533. PMLR.

\bibitem[{Zadeh et~al.(2020)Zadeh, Edo, Awad, and Moshovos}]{zadeh2020gobo}
Ali~Hadi Zadeh, Isak Edo, Omar~Mohamed Awad, and Andreas Moshovos. 2020.
\newblock Gobo: Quantizing attention-based nlp models for low latency and
  energy efficient inference.
\newblock In \emph{IEEE/ACM International Symposium on Microarchitecture
  (MICRO)}, pages 811--824. IEEE.

\bibitem[{Zafrir et~al.(2019)Zafrir, Boudoukh, Izsak, and
  Wasserblat}]{zafrir2019q8bert}
Ofir Zafrir, Guy Boudoukh, Peter Izsak, and Moshe Wasserblat. 2019.
\newblock Q8bert: Quantized 8bit bert.
\newblock Preprint arXiv:1910.06188.

\bibitem[{Zhang et~al.(2018)Zhang, Dinan, Urbanek, Szlam, Kiela, and
  Weston}]{zhang2018personalizing}
Saizheng Zhang, Emily Dinan, Jack Urbanek, Arthur Szlam, Douwe Kiela, and Jason
  Weston. 2018.
\newblock Personalizing dialogue agents: I have a dog, do you have pets too?
\newblock In \emph{Annual Meeting of the Association for Computational
  Linguistics (Volume 1: Long Papers)}, pages 2204--2213.

\bibitem[{Zhang et~al.(2020)Zhang, Hou, Yin, Shang, Chen, Jiang, and
  Liu}]{ternarybert}
Wei Zhang, Lu~Hou, Yichun Yin, Lifeng Shang, Xiao Chen, Xin Jiang, and Qun Liu.
  2020.
\newblock Ternarybert: Distillation-aware ultra-low bit bert.
\newblock In \emph{Conference on Empirical Methods in Natural Language
  Processing}.

\end{thebibliography}
\bibliographystyle{acl_natbib}

\clearpage
\appendix



\section{Derivation of Gradient of Dynamic Scaling}
\label{apdx:scaling_grad}
In this section,
we provide the derivation of the gradient of the proposed dynamic scaling $\gamma$.
The quantization in the forward process can be written as
\begin{eqnarray*}
    \alpha &=& \frac{\|\m w\|_1}{n} \gamma,\\
\u&=&\text{clip}(\m w, -\alpha, +\alpha)/\alpha,\\
\m w_q&=&Q(\u)\alpha,
\end{eqnarray*}
where $Q(\cdot)$ is a uniform quantization function as described  in Section~\ref{sec:difficulty}.
Based on the chain rule, the gradient of scaling $\gamma$ w.r.t. the training loss function $\ell$ is:
\begin{eqnarray}
\small
\frac{\partial \ell}{\partial \gamma}\!\!\!\!\!& =&\!\! \!\!\!
\sum_{i=1}^n\!\frac{\partial \ell}{\partial [\m w_q]_i}\frac{\partial [\m w_q]_i}{\partial \gamma} \nonumber\\
\!\!\!\!\!& =&\!\!\!\!\! \sum_{i=1}^n\!\frac{\partial \ell}{\partial [\m w_q]_i}(\frac{\partial [\m w_q]_i}{\partial Q(u_i)}\frac{\partial Q(u_i)}{\partial \alpha} \frac{\partial \alpha}{\partial \gamma} \!+\! \frac{\partial [\m w_q]_i}{\partial \alpha}\frac{\partial \alpha}{\partial \gamma}) \nonumber\\
\!\!\!\!\!&=&\!\!\!\!\! \sum_{i=1}^n\! \frac{\partial \ell}{\partial [\m w_q]_i}(\alpha \frac{\partial Q(u_i)}{\partial \alpha}\frac{\|\m w\|_1}{n} \!+\! Q(u_i)\frac{\|\m w\|_1}{n} ) \nonumber\\
\!\!\!\!\!&=&\!\!\!\!\!\sum_{i=1}^n \frac{\partial \ell}{\partial [\m w_q]_i}(\alpha \frac{\partial Q(u_i)}{\partial \alpha} \!+\! Q(u_i)) \frac{\|\m w\|_1}{n}.  \label{eq-scaling}
\end{eqnarray}

We use straight through estimator (STE) to estimate the gradient of uniform quantizer $Q(\cdot)$, 
\textit{i.e.}, $\forall i, \frac{\partial Q(u_i)}{\partial u_i}=1$.
Thus the gradient 
$\frac{\partial Q(u_i)}{\partial \alpha}$
can be written as:
\begin{equation}
\begin{aligned}
\!\! \!\!\frac{\partial Q(u_i)}{\partial \alpha}  \!\!=\!\! \frac{\partial Q(u_i)}{\partial u_i}\frac{\partial u_i}{\partial \alpha}
\!\!=\!\!
 \left\{\begin{matrix}
\!\!\!\!\!\! & 0, w_i \leq -\alpha\\ 
\!\!\!\!\!\! & -\frac{w_i}{\alpha^2}, -\alpha\!<\! w_i \!<\!\alpha \\ 
\!\!\!\!\!\! & 0, w_i \geq \alpha
\end{matrix}\right.
 \end{aligned}.
 \label{eq-alpha}
\end{equation}

By combining Eq. \eqref{eq-scaling} and Eq. \eqref{eq-alpha}, we get


\begin{equation*}
\begin{aligned}
&\frac{\partial \ell}{\partial [\m w_q]_i}\left(\alpha \frac{\partial Q(u_i)}{\partial \alpha} + Q(u_i) \right) \frac{\|\m w\|_1}{n} \\
 &= \left\{\begin{matrix}
 & \frac{\partial \ell}{\partial [\m w_q]_i}Q(u_i) \frac{\|\m w\|_1}{n}, w_i\leq -\alpha\\ 
 & \frac{\partial \ell}{\partial [\m w_q]_i}[- \frac{w_i}{\alpha} + Q(u_i) ] \frac{\|\m w\|_1}{n}, -\alpha<w_i<\alpha \\ 
 & \frac{\partial \ell}{\partial [\m w_q]_i}Q(u_i) \frac{\|\m w\|_1}{n}, w_i\geq \alpha
\end{matrix}\right.
 \end{aligned}
\end{equation*}
which considers both the weight inside and outside the clipping value, and is proportional to the weight magnitude $\frac{\|\m w\|_1}{n}$.

\section{More Experimental Settings}
\subsection{Datasets}
\label{apdx:dataset}
The train/val/test splits for different datasets are
shown on Table \ref{table_datasplit}. 

\begin{table}[htbp]
    \centering
	\scalebox{0.85}{
	    \setlength{\tabcolsep}{2.5mm}{
    \begin{tabular}{c|ccc}
    Dataset & Training & Validation & Test\\
    \toprule
     WikiText2&36,717	&3,760&	4,358\\
    PTB	& 42,068&	3,370	&3,761\\	
    WikiText103&1,801,350&	3,760&	4,358\\	
    Persona-Chat&	8,939&	1,000&	968\\	
    XSum	&204,045	&11,332	&11,334\\
    \hline
    \end{tabular}}}
    \caption{
    Data splits of different datasets.}
    \label{table_datasplit}
\end{table}

\subsection{Model Architectures}
\label{apdx:model_arch}

\paragraph{GPT-2.}
The vocabulary size of GPT-2 is 50527.  
We use GPT-2-small with  12 decoder layers and hidden state dimension of 768, for experiments in Sections~\ref{expt:difficulty},~\ref{sec:expt} and ~\ref{sec:discussion}.
GeLU \citep{hendrycks2016gaussian} is used as the activation function.
In the experiments of Appendix \ref{section-larger}, we adopt GPT-2-base with 24 decoder layers and hidden state dimension of 1024, to evaluate the quantization ability on larger models.  

\paragraph{BART.}

The vocabulary size of BART is 50265.
We use BART-base  with 6 encoder layers, 6 decoder layers and hidden state dimension as 768 for experiments in Section~\ref{sec:expt}.
In the experiments of Appendix \ref{section-larger}, we adopt BART-large with 12 encoder layers, 12 decoder layers and hidden state dimension 1024, to evaluate the quantization ability on larger models.  


\setlength{\tabcolsep}{1.3mm}
\begin{table*}[thb]
    \centering
	\scalebox{0.85}{
	    \setlength{\tabcolsep}{2.5mm}{
    \begin{tabular}{c|c|c|ccc|c|cccc}
        Method & \tabincell{c}{\#Bits\\(W-E-A)} & \tabincell{c}{Size\\(MB)($\downarrow$)} &  WikiText2  & PTB & WikiText103 & \tabincell{c}{Size\\(MB)($\downarrow$)} & \multicolumn{3}{c}{XSum} \\
    \toprule
    Metric & & & PPL ($\downarrow$) &PPL ($\downarrow$)&PPL ($\downarrow$)& &R1 ($\uparrow$)&R2 ($\uparrow$) & RL ($\uparrow$)  \\
    \hline
    - &   \textit{full-prec.}& 1353.7&12.46 & 12.35 & 12.37 & 1550.0&	45.25&	22.11&	37.07 \\
    \hline\hline
    PACT & 8-8-8& 342.5 & 12.86 & 13.95 & 13.90&  394.8&	43.55&	20.57&	35.55\\
    \textit{Ours} & 8-8-8& 342.5& \textbf{12.53}	&\textbf{12.40}&	\textbf{12.68}
 &394.8&		\textbf{44.34}&	\textbf{21.41}	&\textbf{36.32}\\			

    \hline
    PACT & 4-4-8 &174.0& 16.10	&	14.19	&	18.07  &202.2&	19.45&	3.53&	15.58\\
    \textit{Ours} & 4-4-8&174.0 &  \textbf{13.34}&	\textbf{12.41}&\textbf{	14.12}&202.2&	\textbf{44.18}&	\textbf{21.31}&	\textbf{36.25}\\
    \hline
    PACT & 2-2-8 &89.7&98.74& 68.55& 86.60&106.0& 8.53&	0.93&	7.25\\
    \textit{Ours} & 2-2-8 &89.7&\textbf{14.53} &\textbf{13.22}&\textbf{14.52}&106.0&	\textbf{42.38}&	\textbf{19.75}&	\textbf{34.57}\\
    \end{tabular}
    }}

    \caption{Ablation study on larger models. We report the results on 24-layer GPT-2 and 24-layer BART.}
    \label{table-large-model}
   
\end{table*}

\subsection{Hyperparameters}
\label{apdx:hyperparam}
\paragraph{Language Modeling.}
The sequence length is 512.  
The learning rate is initialized to 0.0005 (resp. 0.001) for the GPT-2 backbone parameters  (resp. clipping factor $\gamma$) and then linearly decays to 0.
The number of negative samples in each sequence is 64 for the PTB dataset, 
and 32 for the WikiText2 and WikiText103. 
The temperature $\tau$ and momentum coefficient $m$ is 0.1 and 0.5  respectively.
We train with the AdamW  optimizer \cite{loshchilov2017decoupled} with batch size 32. The training epochs for WikiText2, PTB and WikiText103 are set as 80, 120, 8, respectively.

\paragraph{Next Utterance Prediction.}
The sequence length is 512.  
The learning rate is initialized to 0.0005 (resp. 0.001) for the GPT-2 backbone parameters  (resp. clipping factor $\gamma$) and then linearly decays to 0.
The number of negative samples in each sequence is 32. 
The temperature $\tau$ and momentum coefficient $m$ is 0.1 and 0.5, respectively.
We train with the AdamW  optimizer with batch size 16, for a total of 2 epochs.

\paragraph{Abstractive Summarization.}
We set the length of the source sequence (articles) as 512, and pad the target sequence (summaries) to maximum length.  
We use beam search to generate summaries, with  beam size  6 and length penalty 1. 
The learning rate is initialized to 0.0002 (resp. 0.001) for the BART backbone parameters  (resp. clipping factor $\gamma$) and then linearly decays to 0. 
The number of negative samples is 32.
The temperature $\tau$ and momentum coefficient $m$ is 0.1 and 0.5, respectively.
We train with the AdamW  optimizer with batch size 128, for a total of 8 epochs.



\subsection{Description of the Compared Methods}
\label{apdx:q-method}
\paragraph{PACT.}
PACT \citep{choi2018pact} learns a learnable clipping factor for each module 
by gradient descent.
To make the quantization more accurate, 
we adopt a flexible variant of the original PACT, 
with different positive and negative clipping factors $[-\alpha_{neg}, \alpha_{pos}]$, where both $\alpha_{neg}$ and $\alpha_{pos}$ are initialized as 2.5.
\paragraph{LSQ.} LSQ~\citep{esser2020learned} learns the step-size of quantizer for each module by gradient descent.
We use the recommended initialization strategy of the step size as~\citep{esser2020learned}.

\paragraph{LAQ.}
LAQ~\citep{hou2017loss,hou2018loss} is a loss-aware quantization method that views quantization as an optimization problem and solve it via proximal Newton algorithm.  
We use the approximate solver in \citep{hou2018loss} to compute the quantized weights before each forward propagation.

For the self-implemented methods PACT, LSQ and LAQ, we adopt the commonly-used distillation loss adopted in \citep{hinton2015distilling, jiao2020tinybert}. 
Note that these methods are only used for weights and embeddings, while the activations of these methods follow the same setting as our proposed method in Section~\ref{sec:quantization}.
We also tried using the original language modeling loss w.r.t. the ground-truth labels, and distillation loss
over the  attention as \citep{jiao2020tinybert}.
However, these two losses worsens the performance on all three methods.

\subsection{Frameworks of Double-head GPT-2 and BART }
\begin{figure*}[thb]
\centering 
\includegraphics[width=0.9\textwidth]{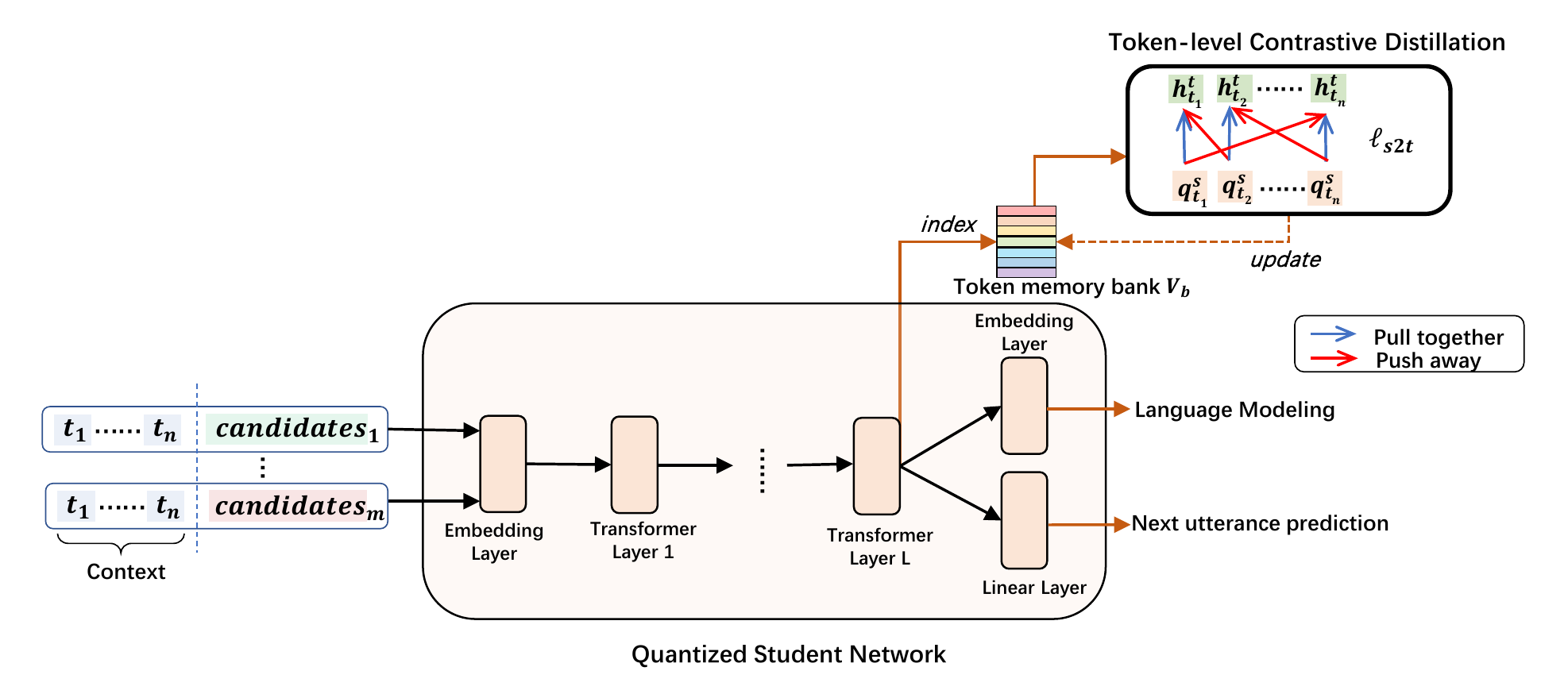}
\caption{
The training workflow of the proposed method for double-head GPT-2 quantization in the task of next utterance prediction. 
The model is trained to find the correct candidate. The  full-precision teacher network and distillation loss $\ell_{dist}$
are omitted for simplicity.
} 
\label{fig:model-gptdouble}
\vspace{-0.1in}
\end{figure*}

\begin{figure*}[thb]
\centering 
\includegraphics[width=0.9\textwidth]{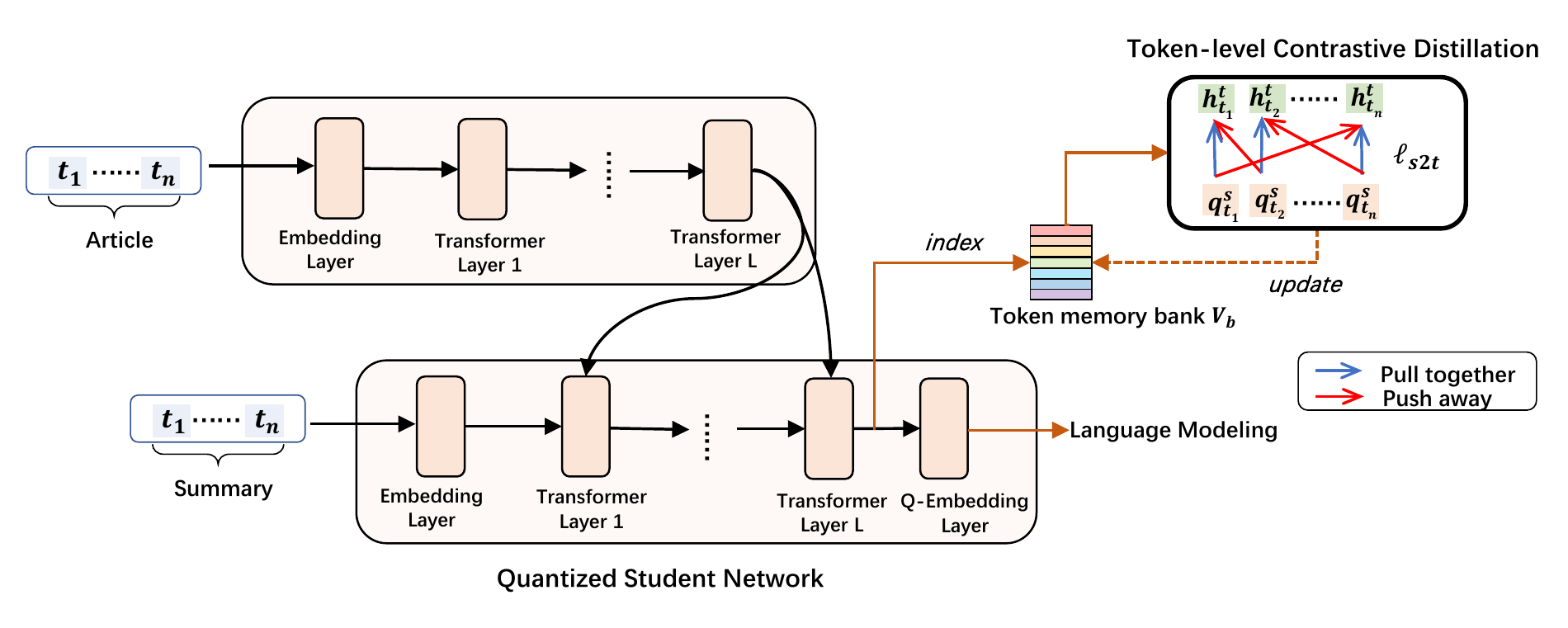}
\caption{The training workflow of the proposed method for BART quantization in the task of abstractive summarization. The  full-precision teacher network and distillation loss
$\ell_{dist}$
are omitted for simplicity. } 
\vspace{-0.1in}
\label{fig:model-bart}
\end{figure*}

Since we adopt double-head GPT-2 and BART for  next utterance prediction and abstractive summarization, the frameworks for these tasks are slightly modified from that on language modeling due to the difference of tasks. 
In Figure \ref{fig:model-gptdouble} and  \ref{fig:model-bart}, we illustrate the framework for  double-head GPT-2 and BART, respectively. 
In the double-head GPT-2, we also quantize the final linear layer in the output head. 

\section{More Experimental Results}
\subsection{Performance of Larger Models}
\label{section-larger}
In Table \ref{table-large-model}, we experiment with GPT-base and BART-large, which both have 24 Transformer layers. 
For all bit-widths,
the training of our method converges successfully without gradient exploding/vanishing problems.  
QuantGPT outperforms PACT by a large margin in all tasks, 
especially for 2-bit weight. 
Our quantization method 
on larger models also
has better performance than that on 
12-layer GPT-2 and 12-layer BART in Section~\ref{sec:expt}.

\begin{figure*}[t]
    \centering
	\subfigure[ "there is no asbestos in our products now"]{
		\includegraphics[width=0.95\textwidth]{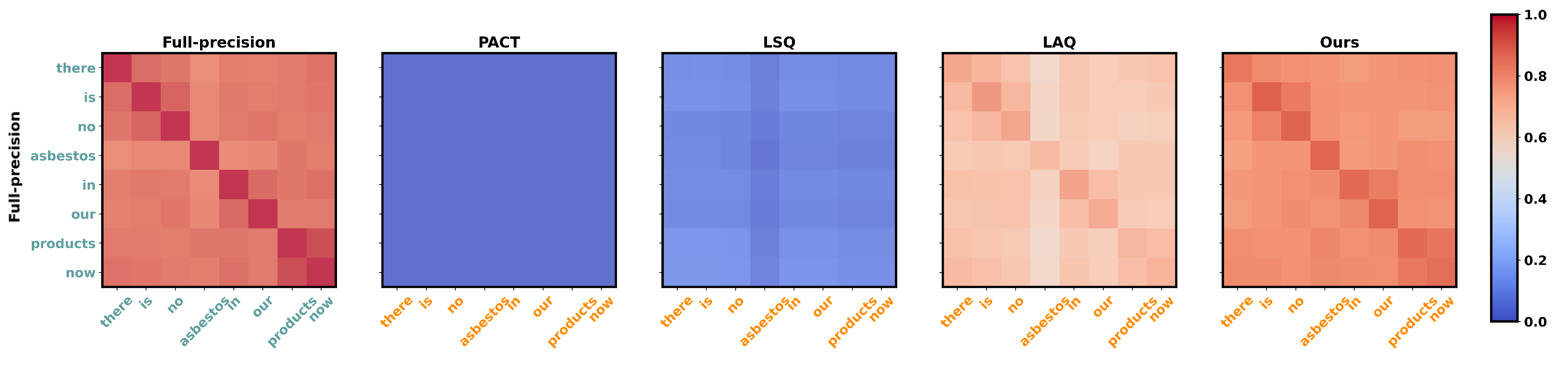}
		\label{fig:vis1}
	}
	
	\subfigure["cray computer has applied to trade on nasdaq"]{
		\includegraphics[width=0.95\textwidth]{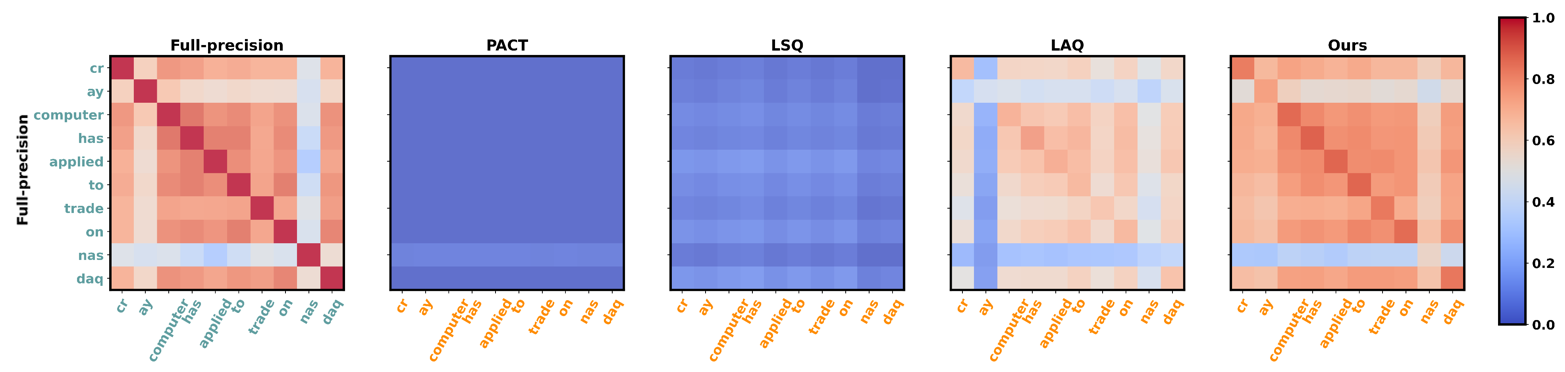}
		\label{fig:vis2}
	}

	\subfigure["no price for the new shares has been set"]{
		\includegraphics[width=0.95\textwidth]{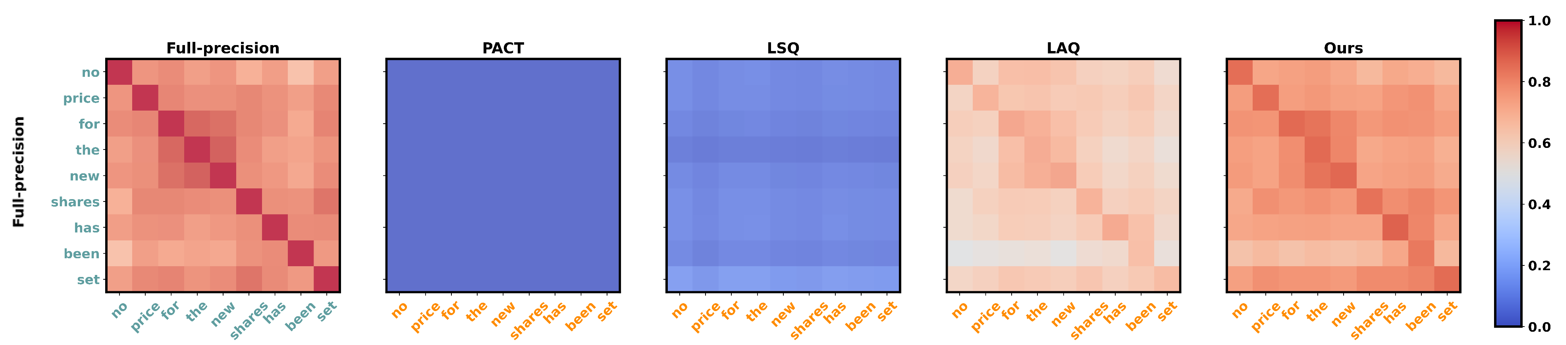}
		\label{fig:vis3}
    }
    
	\subfigure["the centers normally are closed through the weekend"]{
		\includegraphics[width=0.95\textwidth]{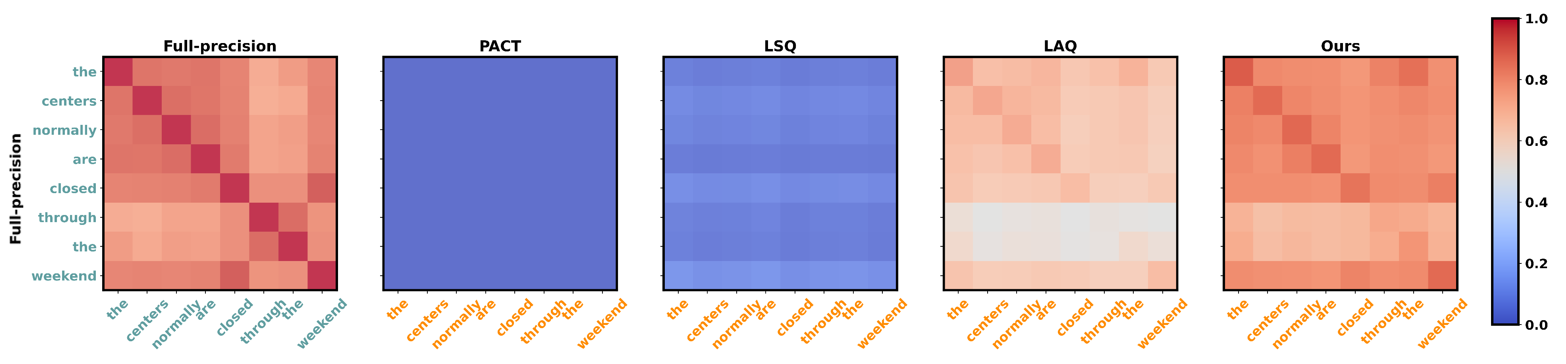}
		\label{fig:vis4}		
	}
\caption{More Visualizations: matrices representing the cosine similarities between representations of all pairs of tokens in a sentence,
between the full-precision model and 2-bit quantized models trained on PTB dataset.  
Token representations at the last decoder layer of GPT-2 are used.}
\label{fig_cossim_more}
\end{figure*}

\subsection{Examples of Summarizations}
\label{apdx:example_summarization}
In  Table \ref{table-sum-demo}, we provide the example summarizations on the XSum dataset. 
By comparing the articles, references and generations, 
the generated summaries by our quantized model are more logical and terse
than
PACT, 
LSQ and LAQ, 
which face problems of homogeneous word embeddings to some extent as discussed in Section~\ref{sec:difficulty}. 
\label{apdx:demo}
\begin{table*}

    \centering
    \small
    \renewcommand\arraystretch{1.4}{
    \begin{tabular}{p{0.95\textwidth}}
    
    \textbf{Article:} On Tuesday, a BBC Spotlight programme revealed that eight children had gone missing in Northern Ireland. Two of the girls were Somali teenagers who disappeared in 2005 and 2012. The Health and Social Care Board has said new guidelines are in place and add that no children have gone missing since 2014. Separated children are children outside their country of origin and separated from their parents or legal guardian. The term can also include unaccompanied asylum-seeking children and trafficked children. When they arrive in Northern Ireland they are taken into the care of the local health trust. Eight children have gone missing since 2005 and they remain missing. The SDLP's health spokesman Mark H Durkan said he would be raising the issue at the Northern Ireland Assembly's health committee and his party colleague Alex Attwood would be raising it at the justice committee. "The number of children who cannot be accounted for is something that needs urgent inquiry and investigation," he said. "There is a lot of very good work being done to look after the welfare of unaccompanied young people, but clearly we now have some very big questions that need to be answered." Ulster Unionist MLA Jo-Anne Dobson said it was "frankly appalling" to hear that eight children had gone missing. "I have written to Health Minister Michelle O'Neill on this issue to seek further clarification and to demand details of how the department, health trusts and the Health and Social Care Board have sought to address each of the cases involved in the investigation," she added. The Green Party leader Steven Agnew also said it was extremely worrying that children can disappear without a trace. Paula Bradshaw from the Alliance Party added that the health trusts and police "need to work closer over the handling of these cases". In a statement, the Police Ombudsman for Northern Ireland said: "Our director of investigations will be reviewing the contents of the programme to ascertain if there are any issues of police conduct which may need further investigation." The Police Service of Northern Ireland has said that in the two cases identified in the programme, investigations were robust and all information available at the time was followed. The Health and Social Care Board has said that new guidelines are in place and stress that no children have gone missing since 2014. BBC Spotlight's investigation is now available on BBC iPlayer.\\
    \hline
   \textbf{Reference:} An urgent inquiry is needed into separated children who have gone missing from care, the Social Democratic and Labour Party has said.\\
   \hline
    \textbf{PACT:} TheTheAAATheTheTheAnAnAnTheThe an an an been been been jailed. \\
   \hline
    \textbf{LSQ:} The SDLP has called for an urgent inquiry into the welfare of unaccompanied children in Northern Ireland. \\  
    \hline
    \textbf{LAQ:} The SDLP has called for "urgent inquiry and investigation" into the handling of unaccompanied children in Northern Ireland. \\
    \hdashline
    \textbf{Ours:} The SDLP is calling for an urgent inquiry and investigation into the disappearance of unaccompanied young people. \\
    \hline\hline
    \\ \\
    \textbf{Article:} The dairies operation, which processes and distributes milk, is being sold to Germany's Mueller for Â£80m.  It comes as profits at the UK's largest dairy food company fell 95\% to Â£900,000 in the six months to September.  Dairy Crest processes and delivers around 1.3 billion litres of milk a year for retailers and homes.  Dairy Crest said in a statement that the deal was in the best interests of consumers, customers and dairy farmers.  The dairies business accounts for about 70\% of the company's revenues, which rose 1\% to Â£682.1m during the six months.  After the sale, which still needs shareholder approval and could take several months, Dairy Crest will focus on its profitable cheese and spreads operations.  There are about 14,000 dairy farmers in the UK, producing 3.3 million litres a day. However, with milk prices having fallen, there has been much debate about whether the economics of the industry are sustainable.  Investors approved of the Dairy Crest's decision to get out of a loss-making sector, sending its shares 10\% higher in morning trading on Thursday.  Muller said the deal would lead to lower costs and larger exports of dairy products made in the UK.  Ronald Kers, chief executive of Muller UK \& Ireland, said: "We are concerned that the dynamics of the UK fresh milk market are unsustainable for dairy processors in the mid to long term and this acquisition will allow us to reduce our costs, increase our efficiencies and invest in the future."  Under the deal, Mueller's UK division - Muller Wiseman Dairies - will take over factories at Foston, in Derbyshire, Chadwell Heath, in Essex, and Severnside, near Gloucester.  The deal also includes the Hanworth glass bottling site in Middlesex, where Dairy Crest is consulting with employees on the site's future, and 72 depots.  Muller bought Robert Wiseman in 2012.\\
    \hline
   \textbf{Reference:} Dairy Crest, maker of Cathedral City cheese and Country Life butter, has announced a big slump in profits and the sale of its milk business.\\
   \hline
    \textbf{PACT:} More than than more more more than more than than than to be be be will will will be be are are are be be to the. \\
   \hline
    \textbf{LSQ:} Dairy Crest is to sell its Dairies business to a German company for an undisclosed sum. \\  
    \hline
    \textbf{LAQ:} Dairy giant Dairy Crest is to sell its UK business to a German company for an undisclosed sum.\\
    \hdashline
    \textbf{Ours:} Dairy Crest, the world's largest dairy producer, is to sell its UK operations to a German firm.\\
    \hline\hline
    \end{tabular}
    }
    \caption{Example summaries generated by 2-bit BART quantized with different  methods.}
    \label{table-sum-demo}
\end{table*}

\subsection{More Visualizations for the Token Representations}
\label{apdx:vis_token}

In Figure \ref{fig_cossim_more}, we provide the visualizations of token representations on more samples.
The observations are similar to those in Section~\ref{sec:difficulty}.

\end{document}